\documentclass{article} 
\usepackage{iclr2026_conference,times}

\usepackage{amsmath,amsfonts,bm}

\def\eqref#1{equation~\ref{#1}}

\def\1{\bm{1}}

\DeclareMathAlphabet{\mathsfit}{\encodingdefault}{\sfdefault}{m}{sl}
\SetMathAlphabet{\mathsfit}{bold}{\encodingdefault}{\sfdefault}{bx}{n}

\newcommand{\ourmethod}{\textsc{PAGE-4D}} 

\usepackage{hyperref}
\usepackage{url}
\usepackage{makecell}
\usepackage{float} 
\usepackage{amsmath}
\usepackage[dvipsnames]{xcolor}
\definecolor{col1}{RGB}{232, 161, 148}
\definecolor{col2}{RGB}{148, 187, 232}
\usepackage{colortbl}   
\usepackage{bm}
\usepackage{mathrsfs} 
\usepackage{dsfont}
\usepackage{multirow}
\usepackage{graphicx}
\usepackage{natbib}
\usepackage{caption}
\usepackage{amsfonts}
\usepackage{xcolor}
\usepackage{booktabs}
\usepackage{tikz}
\usepackage{xfp}
\usepackage{wrapfig}   
\usepackage{capt-of} 
\usepackage{annotate_equations}

\usepackage{algorithm}
\usepackage{algorithmic}
\usepackage{newfloat}
\usepackage{listings}
\DeclareCaptionStyle{ruled}{labelfont=normalfont,labelsep=colon,strut=off}
\floatstyle{ruled}
\newfloat{listing}{tb}{lst}{}
\floatname{listing}{Listing}

\definecolor{darkgreen}{rgb}{0.0, 0.5, 0.0}

\title{\ourmethod: Disentangled pose and geometry estimation for vggt-4d perception}

\author{Kaichen Zhou$^{1,2}$\quad Yuhan Wang$^{2,3*}$\quad Grace Chen$^{1*}$ \quad Xinhai Chang$^{1,2*}$  \\ \textbf{Gaspard Beaudouin$^{1,4}$\quad Fangneng Zhan$^2$  \quad Paul Pu Liang$^{2\dagger}$\quad Mengyu Wang$^{1,5\dagger}$} \\
$^1$Harvard AI and Robotics Lab, Harvard University \\
$^2$Media Lab and Electrical Engineering and Computer Science, Massachusetts Institute of Technology \\
$^3$Department of Computing, Imperial College London  \\
$^4$École Nationale des Ponts et Chaussées,  Institut Polytechnique de Paris
\\
$^5$Kempner Institute for the Study of Natural and Artificial Intelligence, Harvard University \\
}

\iclrfinalcopy 

\begin{document}
\renewcommand{\thefootnote}{}
\footnotetext{$^*$These authors contributed equally as the second authors. $^\dagger$These authors jointly supervised this work.}
\footnotetext{The acknowledgments for the video used in Fig.~\ref{fig:teaser} and Fig.~\ref{fig:point} are provided in the appendix.}
\maketitle

\begin{figure}[h]
\vspace{-0.5em}
    \includegraphics[width=\textwidth]{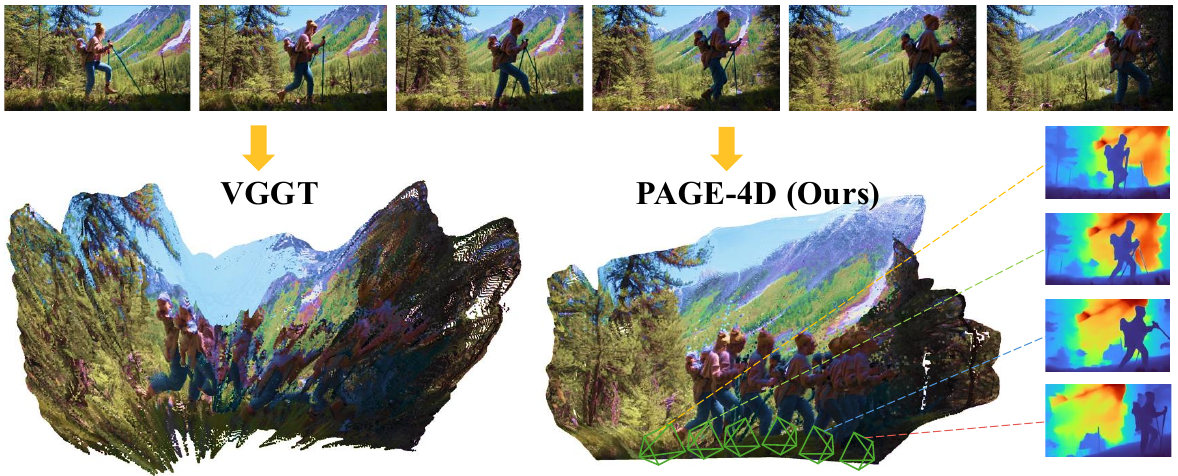}
    \caption{\textbf{\ourmethod} takes a sequence of RGB images depicting a dynamic scene as input and simultaneously predicts the corresponding camera parameters and 3D geometry information—all within a fraction of a second. Compared to VGGT, \textbf{\ourmethod} produces denser and more accurate point cloud reconstructions with better depth estimation quality. (Best viewed in PDF.) 
    }
    \label{fig:teaser}
\end{figure}

\begin{abstract}
Recent 3D feed-forward models, such as the Visual Geometry Grounded Transformer (VGGT), have shown strong capability in inferring 3D attributes of static scenes.
However, since they are typically trained on static datasets, these models often struggle in real-world scenarios involving complex dynamic elements, such as moving humans or deformable objects like umbrellas.
To address this limitation, we introduce \ourmethod, a feedforward model that extends VGGT to dynamic scenes, enabling camera pose estimation, depth prediction and point cloud reconstruction —all without post-processing.
A central challenge in multi-task 4D reconstruction is the inherent conflict between tasks: accurate camera pose estimation requires suppressing dynamic regions, while geometry reconstruction requires modeling them. To resolve this tension, we propose a dynamics-aware aggregator that disentangles static and dynamic information by predicting a dynamics-aware mask—suppressing motion cues for pose estimation while amplifying them for geometry reconstruction.
Extensive experiments show that \ourmethod\ consistently outperforms the original VGGT in dynamic scenarios, achieving superior results in camera pose estimation, monocular and video depth estimation, and dense point map reconstruction. 
Our code is available at \textcolor{red}{\href{https://page4d.github.io}{\texttt{Link}}}, including both the training-and-inference masking variant and the training-only masking variant (= VGGT architecture at inference).

\end{abstract}

\section{Introduction}

Despite recent advances in feedforward 3D estimation of static scenes from image sets~\citep{zhang2025advances,wang2025vggt,wang2024dust3r}, extending these capabilities to dynamic environments—scenes where objects or people undergo motion or deformation— remains a significant challenge due to the complexity of real-world motion. 
A common strategy for handling dynamic scenarios is to decompose the problem into a series of sub-modules, such as depth estimation, optical flow computation, and object tracking~\citep{luiten2020track,mustafa2016temporally,kopf2021robust,zhang2022structure}. While this modular approach simplifies the task by disentangling different components, it often results in increased computational cost and error accumulation across sequential stages~\citep{zhang2025monst3r}.
Given the limitations of modular pipelines, a unified method for dynamic geometry learning that avoids sequential decomposition offers a more effective and coherent solution. However, developing such models usually requires capturing spatiotemporal relationships across frames, and demands notable computational resources as well as access to large-scale dynamic datasets with ground-truth geometry \citep{zhang2025monst3r,wang2025continuous}. 

Motivated by these challenges, we present \textbf{\ourmethod}, a unified and efficient feed-forward model that enables the inference of key 3D attributes in dynamic scenes as shown in Fig.\ref{fig:teaser}. To address the limited availability of labeled dynamic data, we build on the pretrained 3D foundation model VGGT~\citep{wang2025vggt} and adapt it to dynamic scenarios through targeted fine-tuning.
While VGGT demonstrates strong performance in static scene understanding, its accuracy drops significantly when applied to dynamic environments involving people, vehicles, or deformable objects. This limitation stems from a fundamental tension: motion provides valuable cues for geometry estimation in dynamic scenarios, yet simultaneously introduces noise that corrupts camera pose estimation by violating the static epipolar constraint, as shown in Fig.~\ref{fig:motivation}~(a).
In other words, the very signals that enable reconstructing dynamic objects are also those that hinder reliable pose recovery~\citep{chen2025easi3r,zhang2025monst3r,zhang2022structure}.

This insight motivates our central idea: rather than viewing dynamics as uniformly harmful or helpful, we disentangle their effects across tasks. We introduce a dynamics-aware aggregator that first predicts a mask to identify dynamic regions, and then applies it via a cross-attention mechanism—filtering dynamic content for camera pose tokens while emphasizing it for geometry tokens. Together with targeted fine-tuning of layers most sensitive to dynamics, this design allows us to harness motion where it benefits geometry grounding, while suppressing its negative impact on pose estimation. With this design, our method achieves accurate pose and geometry estimation for both static and dynamic content in challenging dynamic scenarios, as illustrated in Fig.~\ref{fig:teaser}.
Through extensive experiments, \ourmethod\ establishes new state-of-the-art performance across multiple benchmarks and tasks. For instance, on the Sintel benchmark, it reduces the camera pose estimation ATE from 0.214 (VGGT) to 0.143 and improves the scale-aligned video depth Abs Rel from 0.484 to 0.357. Notably, thanks to its plug-in design, \ourmethod{} adds only a negligible overhead in both runtime and storage compared to VGGT. This work makes the following key contributions:
\begin{itemize}
    \item We propose \ourmethod, a dynamic-aware extension of VGGT for 4D scene understanding, which achieves state-of-the-art results on dynamic geometry perception benchmarks.
    \item We design a dynamics-aware aggregator that combines (i) a mask prediction module for identifying dynamic regions and (ii) a global attention mechanism that selectively leverages or suppresses dynamic information across tasks.
    \item We provide an in-depth analysis of VGGT under dynamic conditions and introduce a targeted fine-tuning strategy that adapts only the layers most sensitive to dynamics, enabling efficient transfer by updating only a limited subset of parameters.

    \item Finally, we provide both inference-time and training-only masking variants with comparable performance. The latter removes the dynamic mask during inference and therefore preserves the VGGT architecture. This shows that dynamic masking can serve as an effective training-time regularizer for pose–geometry disentanglement without introducing additional inference-time overhead.
\end{itemize}

\section{Related Work}

\textbf{3D Feedforward Model} is learning-based approach that reconstructs static 3D scene geometry from input images with temporal invariance assumption~\citep{bochkovskii2024depth,yin2023metric3d,piccinelli2024unidepth,leroy2024grounding}, treating all views as capturing the same static scene. DUSt3R~\citep{wang2024dust3r} is the representative of this reconstruction framework, introducing transformer-based architectures that processes image pairs from different viewpoints, learning direct mappings from 2D image pixels to 3D coordinate fields. 
Subsequent works~\citep{tang2024mvdust3rsinglestagescenereconstruction,Yang_2025_Fast3R,wang2025continuous,bhat2023zoedepthzeroshottransfercombining,tang2019banetdensebundleadjustment,Yao_2018_ECCV,chen2021mvsnerf} have explored broader scenarios. Among those, VGGT~\citep{wang2025vggt} presents a unified architecture using alternating attention mechanisms within each frame and across the entire sequence, responding the need of joint prediction of camera poses, depth maps, and point correspondences through integrated training. Despite these advances, traditional 3D methods remain temporally invariant and struggle with dynamic scenes, motivating the need for 4D feedforward approaches that explicitly capture scene dynamics.

\noindent

\textbf{4D Feedforward Model} emerges to reconstruct dynamic scenes by capturing geometric evolution over time from image sequences~\citep{tian2023mononerf,van2022revealing,busching2024flowibr,liang2024feed,zhao2023pseudo}. However, it faces significant challenges in modeling temporal geometry changes, as moving objects violate the rigid geometry assumptions of static methods~\citep{Oliensis2000,ozyesil2017surveystructuremotion,cao2025reconstructing}.
Given DUSt3R's success in static reconstruction, several works~\citep{lu2024align3r,point3r,wang2024spann3r,xu2024das3r,Yao_2025_CVPR} have adapted this framework for dynamic scenarios. While MONST3R~\citep{zhang2025monst3r} fine tunes DUSt3R on video sequences, D²USt3R~\citep{han2025d2ust3r} introduces explicit temporal modeling through 4D pointmap representations and cross-frame attention mechanisms, improving in establishment of correspondences between moving objects across frames. Other efforts include training-free methods like Easi3R~\citep{chen2025easi3r}. Despite the progress shown by these DUSt3R-based approaches in dynamic content, they are all constrained by the pairwise progressing framework in DUSt3R. Alternative approaches~\citep{st4rtrack2025,li2024megasam,xu2025geometrycrafterconsistentgeometryestimation,jiang2025geo4d,jin2025stereo4d,piccinelli2024unidepth,bochkovskii2024depth} explore different architectural designs to handle dynamic scenes for task-specific solutions, but they often sacrifice the generalizability of feedforward approaches. 
More recently, the success of VGGT in sequence-based static reconstruction has inspired its extensions~\citep{li2025light} to dynamic scenarios. 
Recent approaches such as MoVieS~\citep{lin2025movies} and StreamVGGT~\citep{zhuo2025streamvggt} focus on narrow, application-specific scenarios rather than the broader challenge of adapting static models to dynamic domains. In contrast, we propose \ourmethod{} to address this general challenge, demonstrating that carefully targeted fine-tuning of key attention components can effectively bridge the static–dynamic divide without requiring major architectural changes.

\section{Methodology}
In this paper, we extend the VGGT to \ourmethod{} (Disentangled Pose and Geometry Estimation for 4D Perception), a dynamic-aware framework for robust 4D scene understanding. Given a sequence of $N$ RGB frames $\{\mathbf{I}_i\}_{i=1}^{N}$ captured in a dynamic environment, our objective is to predict temporally consistent 3D outputs for each frame:

\begin{equation*}
    f\bigl(\{\mathbf{I}_i\}\bigr) = \bigl\{(\mathbf{g}_i,\, \mathbf{D}_i,\, \mathbf{P}_i\bigr\}_{i=1}^{N},
\end{equation*}
where $\mathbf{g}_i \in \mathbb{R}^9$ encodes the camera intrinsics and extrinsics, $\mathbf{D}_i \in \mathbb{R}^{H \times W}$ is the depth map, and $\mathbf{P}_i$ the 3D point map.  

In this section, we begin by examining the behavior of VGGT under dynamic conditions and analyzing how its transformer architecture represents spatiotemporal information. This analysis reveals fundamental limitations when directly applying VGGT to dynamic scenes. Guided by these insights, we introduce \ourmethod{}, a principled yet lightweight extension of VGGT that enables accurate and efficient estimation of camera pose, geometry, and tracking in challenging dynamic environments.

\subsection{Motivation}
\textbf{Empirical Observation:} Although VGGT achieves state-of-the-art performance in static scene understanding, its accuracy degrades markedly in the presence of dynamic objects. 
On the Odyssey test set~\citep{zheng2023pointodyssey}, which evaluates long-range point tracking and geometry understanding in dynamic scenes, we directly apply VGGT for evaluation. The results reveal a clear gap between static and dynamic regions: the Absolute Depth Error in dynamic regions is 94\% higher than in static regions. 
These results highlight the need for an architecture that achieves reliable scene understanding across both static and dynamic scenarios.

\begin{figure*}[t]
    \centering
    \includegraphics[width=0.9\textwidth]{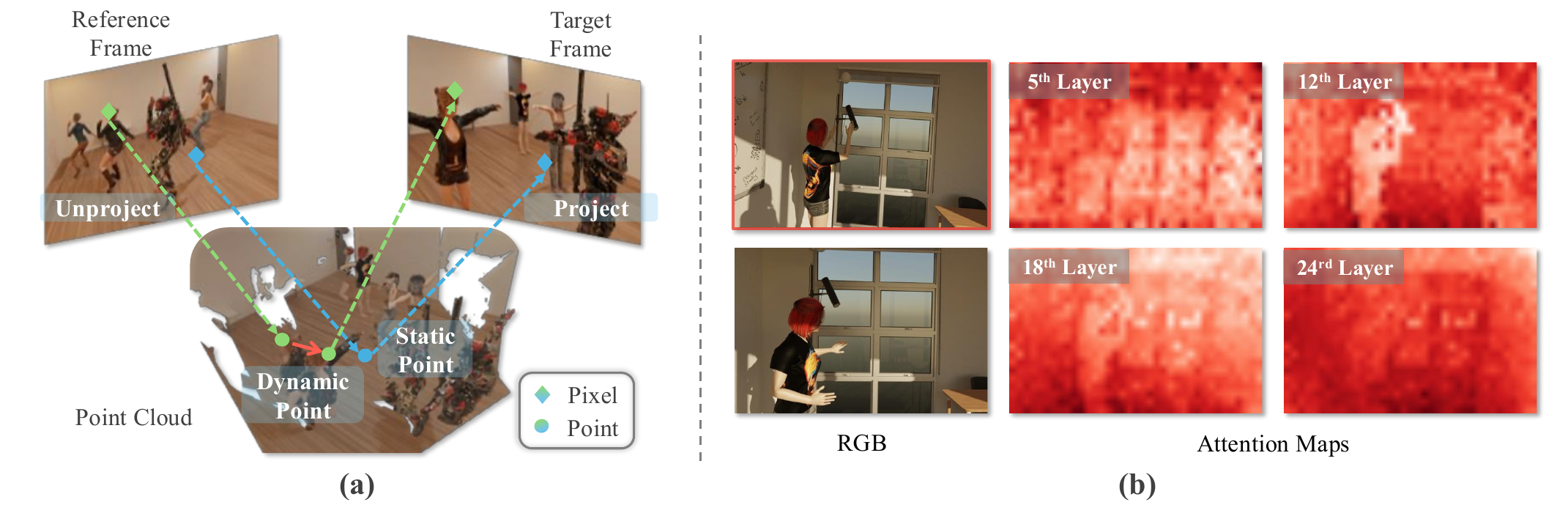}
    \caption{\textbf{Motivating illustration:} (a) In static scenes, geometric consistency is preserved across frames, while in dynamic scenes, moving objects violate this consistency. (b) Visualization of VGGT attention maps from the 5th, 12th, 18th, and 24th layers of global attention block with the method in \cite{caron2021emerging}.
    Attention values are visualized using a white-to-red color map, with white indicating low values and red indicating high values. VGGT tends to ignore dynamic content during the feed-forward process, which motivates our design of the dynamics-aware mask.}
    \label{fig:motivation}
\end{figure*}

To better understand this gap, we first follow \cite{chefer2021transformer} on feature visualization and examine key layers of VGGT (Fig.~\ref{fig:motivation} (b)). We observe that dynamic regions exhibit weaker activations compared to static ones, suggesting that VGGT tends to ignore dynamic content. We then perform an ablation in which attention among dynamic tokens is explicitly suppressed (see Appendix). Masking dynamic patches from the cross-frame attention mechanism improves camera pose estimation, but at the same time leads to a sharp drop in geometry. 
Together, these findings reveal a fundamental tension in dynamic scenes: 
\emph{while camera pose estimation benefits from suppressing dynamic regions to maintain epipolar consistency, 
geometry requires exploiting their motion cues}. 

\textbf{Static Case – Geometric Foundations:} Formally, under static conditions, \underline{geometry estimation} can be achieved by implicitly modeling the correspondence between a reference-frame homogeneous pixel $\mathbf{x}_r$ and its target-frame homogeneous pixel $\mathbf{x}_t$ (Fig.~\ref{fig:motivation} (a))~\citep{zhang2025monst3r,chen2025easi3r}, which is fully determined by the camera intrinsics, depth, and relative pose:
\begin{figure}[h]
\vspace{1.6em}
\begin{equation}
\label{eqn.homo_stat}
    \eqnmarkbox[Emerald]{xt}{\mathbf{x}_t} = \eqnmarkbox[OliveGreen]{K}{\mathbf{K}} \left[ \eqnmarkbox[NavyBlue]{rtr}{\mathbf{R}_{t \leftarrow r}} \, \eqnmarkbox[WildStrawberry]{dep}{D_r(\mathbf{x}_r)} \, \eqnmarkbox[OliveGreen]{K}{\mathbf{K}^{-1}} \eqnmarkbox[BurntOrange]{xr}{\mathbf{x}_r} + \eqnmarkbox[Plum]{ttr}{\mathbf{t}_{t \leftarrow r}} \right],
\end{equation}
\annotate[yshift=1em]{above,left}{xt}{ Pixel (Target Frame)}
\annotate[yshift=1em]{above,left}{K}{Inversed Intrinsic}
\annotate[yshift=-1em]{below,right}{xr}{Pixel (Reference Frame)}
\annotate[yshift=1em]{above,right}{ttr}{Translation Vector}
\annotate[yshift=-1em]{below,left}{rtr}{Rotation Matrix}
\annotate[yshift=-1em]{below,left}{dep}{Depth}
\vspace{0.1em}
\end{figure}

This equation encodes the standard rigid-scene geometry assumption: once depth and camera motion are known, pixel correspondences across frames can be predicted without ambiguity. Meanwhile, \underline{pose estimation}~\citep{zhang2025monst3r,chen2025easi3r}, due to the concentration of relative camera motion, in VGGT often reduces to fitting an essential matrix $\mathbf{E}$ that enforces the epipolar constraint between normalized homogeneous
pixels $\tilde{\mathbf{x}}_r$ and $\tilde{\mathbf{x}}_t$:
\begin{figure}[h]
\vspace{1.6em}
\hspace{5em} 
\begin{minipage}{0.9\linewidth} 
\begin{equation}
\label{eqn.essen_stat}
    \eqnmarkbox[Emerald]{xt}{\tilde{\mathbf{x}}_{t}^{\top}}
    \eqnmarkbox[OliveGreen]{E}{\mathbf{E}}
    \eqnmarkbox[BurntOrange]{xr}{\tilde{\mathbf{x}}_{r}}
    = 0 , \;
    \eqnmarkbox[OliveGreen]{E}{\mathbf{E}}
    = [\eqnmarkbox[Plum]{ttr}{\mathbf{t}_{t \leftarrow r}}]_\times
    \eqnmarkbox[NavyBlue]{rtr}{\mathbf{R}_{t \leftarrow r}}.
\end{equation}
\annotate[yshift=1em]{above,right}{xt}{Homogeneous Pixel (Target Frame)}
\annotate[yshift=-1em]{below,right}{E}{Essential Matrix}
\annotate[yshift=-1em]{below,left}{xr}{Homogeneous Pixel (Reference Frame)}
\end{minipage}
\vspace{0.1em}
\end{figure}


\emph{Summary:} Under static conditions, both Eqn.~\ref{eqn.homo_stat} and Eqn.~\ref{eqn.essen_stat} hold for all non-occluded pixel pairs $(\mathbf{x}_{r}, \mathbf{x}_{t})$ between the reference and target frames. This makes the joint optimization of camera tokens and geometry tokens over the same patches within a frame a reasonable design choice.


\textbf{Dynamic Case – Violation and Residuals:} In dynamic scenes, to realize \underline{geometry estimation}, motion information needs to be taken into consideration for accurate prediction:
\begin{equation}
\mathbf{x}_t = \mathbf{K} \left[ \mathbf{R}_{t \leftarrow r} \, D_r(\mathbf{x}_r) \, \mathbf{K}^{-1} \mathbf{x}_r + \mathbf{t}_{t \leftarrow r} \right] 
+ \mathbf{K} \mathbf{M}_{t \leftarrow r},
\label{eqn.homo2}
\end{equation}

where $\mathbf{M}_{t \leftarrow r}$ represents the displacement induced by object motion. 
Meanwhile, in the presence of dynamic motion, the Eqn.~\ref{eqn.essen_stat} for \underline{pose estimation} no longer holds. The violation manifests as a residual:
\begin{equation}
\delta(\mathbf{x}_{r}) \equiv \tilde{\mathbf{x}}_{t}^{\top} \mathbf{E} \tilde{\mathbf{x}}_{r}
\;\; \approx \;\; \frac{1}{Z_{r}} \, \mathbf{n}(\mathbf{x}_{r})^{\top} \, \Delta \mathbf{X}_{\perp}(\mathbf{x}_{r}),
\label{eqn.essen2}
\end{equation}



where $\mathbf{n}(\mathbf{x}_{r})$ is the unit normal of the epipolar line associated with $\mathbf{x}_r$, and $\Delta \mathbf{X}_{\perp}(\mathbf{x}_{r})$ is the component of the dynamic displacement perpendicular to that line. This residual quantifies the degree to which dynamic motion “pushes” correspondences away from the epipolar geometry predicted by the camera. The larger the residual, the stronger the violation of the static-scene assumption, and the greater the resulting pose estimation error. Eqn.~\ref{eqn.essen2} implies that in dynamic scenarios, only the static subset of pixel pairs $(\mathbf{x}_{r}^{sta}, \mathbf{x}_{t}^{sta})$ satisfy Eqn.~\ref{eqn.essen_stat}.

\emph{Summary:} Under dynamic conditions, Eqn.~\ref{eqn.homo2} for geometry estimation remains valid for all non-occluded pixel pairs $(\mathbf{x}_{r}, \mathbf{x}_{t})$ between the target and reference frames, whereas Eqn.~\ref{eqn.essen_stat} for pose estimation holds only for the static subset of non-occluded pixels $(\mathbf{x}_{r}^{sta}, \mathbf{x}_{t}^{sta})$, as explained in Eqn.~\ref{eqn.essen2}.

\emph{Insight:} Under dynamic scenarios, camera pose estimation is brittle to dynamic motion, as small residuals can corrupt essential matrix fitting, while geometry and tracking tasks can in fact benefit from modeling $\mathbf{M}_{t \leftarrow r}$. Motivated by this insight, we propose \ourmethod{}, a dynamic-aware extension of VGGT that disentangles the role of dynamic regions across tasks—suppressing them for pose estimation while leveraging them for geometry and tracking.

\begin{figure*}[t!]
    \centering
    \vspace{-0.1cm}
    \includegraphics[width=1.0\textwidth]{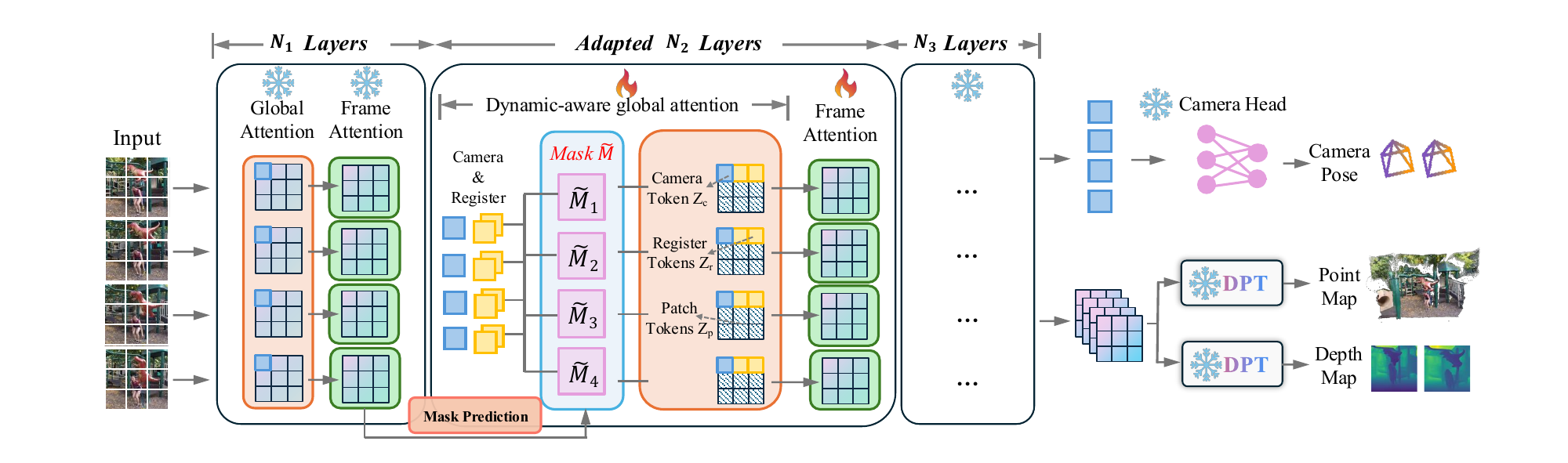}
    \caption{\textbf{Fine-tuning strategy:} 
    Instead of fine-tuning the entire VGGT architecture, we adapt only the middle $N_2$ layers of the global attention mechanism, which are most critical for cross-frame information fusion. To further address dynamic scenes, we introduce a \textit{dynamics-aware aggregator} that predicts a mask to disentangle dynamic and static content. \textbf{In our code}, we provide both the training-and-inference masking variant and the training-only masking variant (= VGGT architecture at inference).}
    \label{fig:framework}
\end{figure*}

\subsection{\ourmethod{}}

\ourmethod{} is composed of four key components:  
(1) a pre-trained DINO-style~\citep{zhang2022dino} encoder that extracts image-level representations;  
(2) a \textit{dynamics-aware aggregator} that integrates spatial and temporal cues through three modules—Frame Attention for inter-frame patch relations, Global Attention for intra-frame patch relations, and Dynamics-Aware Global Attention for disentangling dynamic from static content;  
(3) lightweight decoders for depth, 3D point maps; and  
(4) a larger decoder dedicated to camera pose estimation.  

\ourmethod{} inherits components (1), (3), and (4) directly from VGGT, while extending component (2) into a three-stage dynamics-aware aggregator as in Fig.~\ref{fig:framework}. 
The first stage consists of $N_1$ layers, each composed of one Global Attention and one Frame Attention block. Its output is fed into a dynamic mask prediction module, which produces a dynamics-aware mask. 
This mask is then applied in the second stage to disentangle dynamic and static content for pose and geometry estimation. The second stage itself consists of $N_2$ layers, each comprising a Dynamics-Aware Global Attention block and a Frame Attention block.
The final stage consists of $N_3$ layers as the first stage. 

\subsubsection{Dynamics Mask Prediction}

A central challenge in dynamic scenes is to selectively suppress the influence of moving objects for tasks such as pose estimation, while still retaining their information for geometry. To achieve this, we design a dynamic mask prediction module that learns, in a self-supervised manner, which spatial regions are likely to correspond to dynamic objects. 
This is feasible because the middle layers of \ourmethod{} already disentangle dynamic and static content, as illustrated in Fig.~\ref{fig:motivation}(b), where dynamic regions are treated distinctly.
Formally, given the token features $\mathbf{z} \in \mathbb{R}^{B \times S \times P \times d}$ from the aggregator, we first extract only the patch tokens $\mathbf{z}_p \in \mathbb{R}^{B \times S \times (H_P \cdot W_P) \times d}$. These tokens are projected into a lower-dimensional representation via a linear mapping, followed by a depthwise convolutional head that produces mask logits:
$
\mathbf{m} = \mathrm{Conv}\bigl(\mathbf{z}_p\bigr) \in \mathbb{R}^{(B \cdot S) \times 1 \times H_P \times W_P}.
$

\begin{figure*}[t!]
    \includegraphics[width=\linewidth, clip]{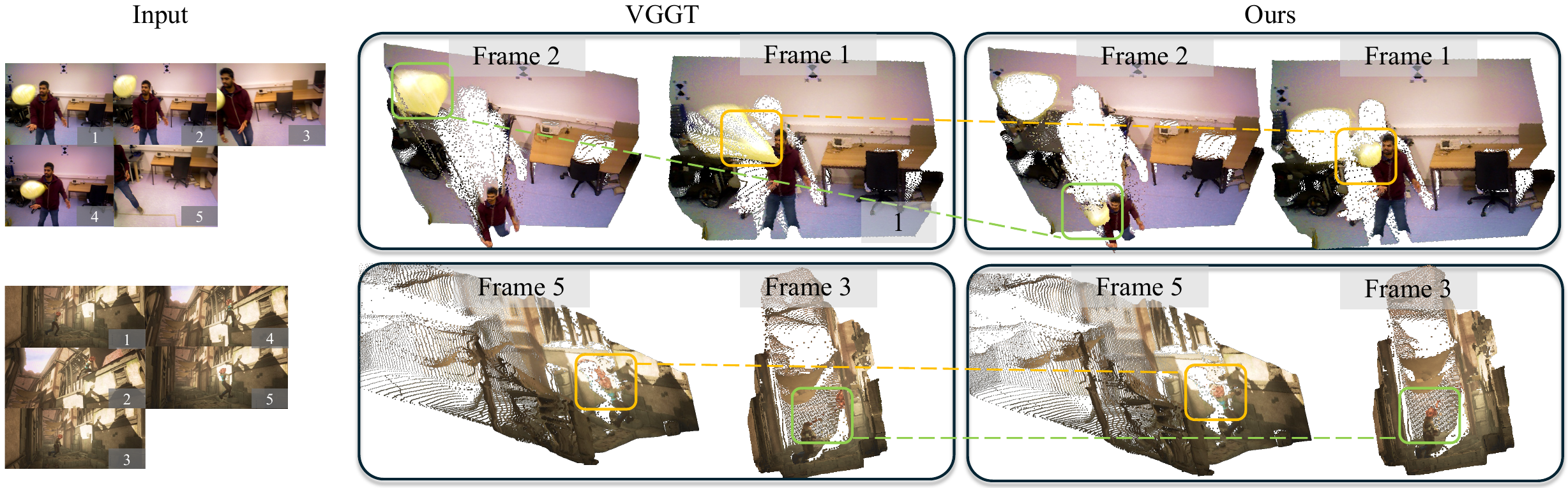}   
    \caption{\textbf{Qualitative Comparison of Point Cloud Estimation on the Bonn \& Sintel}: As shown in the figure, our method effectively captures the geometric structure in scenarios with complex motion, whereas VGGT produces fragmented and inconsistent geometry. (Best viewed in PDF.)}
    \label{fig:Depth1_BS} 
\end{figure*}

\subsubsection{Mask Attention}

Once the dynamic mask $\widetilde{\mathbf{M}}$ has been predicted, it can be directly incorporated into the transformer attention mechanism. Specifically, given queries $\mathbf{Q}$, keys $\mathbf{K}$, and values $\mathbf{V}$, attention is:
\begin{equation}
\mathrm{Attn}(\mathbf{Q},\mathbf{K},\mathbf{V}) = \mathrm{softmax}\!\left(\frac{\mathbf{Q}\mathbf{K}^\top}{\sqrt{d}} + \widetilde{\mathbf{M}}\right)\mathbf{V},
\label{eq:mask}
\end{equation}
where $\widetilde{\mathbf{M}} \in \mathbb{R}^{B \times (S \cdot P) \times (S \cdot P)}$ is the broadcasted mask applied to the attention logits.  
Importantly, we apply this mask in a task-specific manner:  
\textbf{Pose estimation.} For queries corresponding to the camera token and registeration token, $\widetilde{\mathbf{M}}$ actively suppresses attention to dynamic regions, ensuring that pose estimation remains consistent with epipolar geometry and static scene constraints.  
\textbf{Depth and Point Cloud.} For patches concerning these tasks, the mask is not applied, allowing the network to leverage dynamic motion cues to improve point map reconstruction and 2D–3D tracking accuracy.  

This asymmetric design explicitly disentangles the role of dynamic regions across tasks. Dynamic objects, which are detrimental for camera pose estimation, are ignored in that context, but their motion signals remain available for geometry and tracking tasks. By learning the mask in a fully differentiable manner, the model adapts its behavior to the motion patterns present in the training data, rather than relying on pre-defined heuristics.  

\noindent\textbf{Memory-Efficient Mask Mechanism}
Although Eq.~\ref{eq:mask} describes a full $(S\!\cdot\!P)^2$ mask, forming this matrix would require $\mathcal{O}(N^2)$ memory and break fused Scaled Dot-Product Attention, where $N=S\!\cdot\!P$. PAGE-4D instead implements an \emph{equivalent additive mask} using two vectors.
Given attention inputs $\mathbf{Q},\mathbf{K},\mathbf{V} \in \mathbb{R}^{N \times d}$, the mask head predicts:
\[
r \in \mathbb{R}^N,\qquad c \in \mathbb{R}^N.
\]

We append these to the feature dimension:
\[
q'_i = [\, q_i \sqrt{d'/d},\; r_i \sqrt{d'} \,],\qquad
k'_j = [\, k_j,\; c_j \,],\qquad
v'_j = [\, v_j,\; 0 \,],
\]
where $d' = d+1$. Then:
\[
\frac{q'_i k_j'^{\top}}{\sqrt{d'}} =
\frac{q_i^{\top} k_j}{\sqrt{d}} + r_i c_j,
\]
but \emph{without constructing} the $N \times N$ mask.

This uses only $\mathcal{O}(N)$ memory, stays compatible with fused Scaled Dot-Product Attention.

\subsection{Training Details}
\textbf{Fine-tuning Strategy.} During fine-tuning, we update only the middle 10 layers while freezing the remaining aggregator and decoder layers, thereby tuning just 30\% of the model instead of the full network. 
This design is supported by studies on transformer representations, which show that lower layers capture local structures, middle layers model regional relationships, and higher layers encode global semantics~\citep{raghu2021vision,caron2021emerging}. 
Moreover, as illustrated in Fig.\ref{fig:motivation}(b), the middle layers of VGGT tend to suppress dynamic content, leading to degraded performance in dynamic scenarios. By selectively fine-tuning these layers, we aim to reintroduce dynamic information into the feed-forward process. 
Consistent with this intuition, our ablations (Please Refer to Appendix) confirm that the later middle layers contribute most significantly to accurate geometry estimation.


\textbf{Loss Functions.}  
We adopt a multi-task loss combining supervision for camera pose, depth and point-maps:
\begin{equation}
\mathcal{L} = \lambda_c \mathcal{L}_{\text{camera}} + \mathcal{L}_{\text{depth}} + \mathcal{L}_{\text{pmap}}.
\end{equation}
Following VGGT, we empirically set the loss weights to balance gradients across tasks, with $\lambda_c = 5$.
We adopt: Huber loss for camera pose estimation, Uncertainty-weighted depth and point-map losses with gradient regularization.
We do not include point tracking in our model, since the tracking head in VGGT is primarily designed for view registration and is not well-suited to dynamic scenarios. In addition, VGGT does not provide clear training code for the tracking head. These two factors prevent us from incorporating point tracking into our framework. 

\begin{table*}[t]
    \centering
    \caption{
        \textbf{Video Depth Estimation on Sintel~\citep{Butler:ECCV:2012}, Bonn~\citep{palazzolo2019refusion} and DyCheck~\citep{Yang_2025_Fast3R}.} 
        FPS is evaluated on KITTI using one A800 GPU. Missing entries (--) denote results not reported in the original papers cited.}
    \resizebox{\linewidth}{!}
    {
    \begin{tabular}{lccccccccc}
        \toprule
        {\multirow{3}{*}{\textbf{Method}}} &
        {\multirow{3}{*}{\textbf{Params}}} &
        {\multirow{3}{*}{\textbf{Align}}} &
        \multicolumn{2}{c}{\textbf{Sintel}} &
        \multicolumn{2}{c}{\textbf{Bonn}} &
        \multicolumn{2}{c}{\textbf{DyCheck}} &
        \multicolumn{1}{c}{\multirow{3}{*}{\textbf{FPS}}}\\
        \cmidrule(r){4-5} \cmidrule(r){6-7} \cmidrule(r){8-9}
        & & &
        \cellcolor{col1} Abs Rel $\downarrow$ & \cellcolor{col2} $\delta<1.25$ $\uparrow$ &
        \cellcolor{col1} Abs Rel $\downarrow$ & \cellcolor{col2} $\delta<1.25$ $\uparrow$ &
        \cellcolor{col1} Abs Rel $\downarrow$ & \cellcolor{col2} $\delta<1.25$ $\uparrow$ \\
        \midrule
        DUSt3R~\citep{wang2024dust3r} & 571M & \multirow{8}{*}{\makecell{scale \\ Video\\Depth}} & 0.662 & 0.434 & 0.151 & 0.839 & - & - & 1.25\\
        MASt3R~\citep{leroy2024grounding} & 689M & & 0.558 & 0.487 & 0.188 & 0.765 & - & - & 1.01\\
        CUT3R~\citep{wang2025continuous} & 793M & & \underline{0.430} & 0.465 & \textbf{0.077} & \textbf{0.937} & \underline{0.176} & 0.740 & 6.98\\
        Fast3R~\citep{Yang_2025_Fast3R} & 648M & & 0.638 & 0.422 & 0.194 & 0.772 & - & - & 65.8\\
        FLARE~\citep{zhang2025flare} & 1.40B & & 0.729 & 0.336 & 0.152 & 0.790 & - & - & 1.75\\
        VGGT ~\citep{wang2025vggt} & 1.26B & & 0.484 & \underline{0.553} & 0.107 & 0.883 & 0.182 & \underline{0.743} & 43.2\\
        \textbf{\ourmethod (Ours)} & 1.26B & & \textbf{0.357} & \textbf{0.699} & \underline{0.092} & \underline{0.904} & \textbf{0.170} & \textbf{0.785} & 43.2\\
        \midrule
        DUSt3R~\citep{wang2024dust3r} & 571M & \multirow{8}{*}{\makecell{scale\&shift \\ Video\\Depth}}& 0.570 & 0.493 & 0.152 & 0.835 & - & - & 1.25\\
        MASt3R~\citep{leroy2024grounding} & 689M & & 0.480 & 0.517 & 0.189 & 0.771 & - & - & 1.01\\
        CUT3R~\citep{wang2025continuous} & 793M & & 0.534 & 0.558 & \textbf{0.075} & \textbf{0.943} & 0.228 & 0.635 & 6.98\\
        Fast3R~\citep{Yang_2025_Fast3R} & 648M & & 0.518 & 0.486 & 0.196 & 0.768 & - & - & 65.8 \\
        FLARE~\citep{zhang2025flare} & 1.40B & & 0.791 & 0.358 & 0.142 & 0.797 & - & - & 1.75\\
        VGGT ~\citep{wang2025vggt} & 1.26B & & \underline{0.261} & \underline{0.639} & 0.102 & 0.890 & \underline{0.155} & \underline{0.792} & 43.2\\
        \textbf{\ourmethod (Ours)} & 1.26B & & \textbf{0.212} & \textbf{0.763} & \underline{0.090} & \underline{0.903} & \textbf{0.145} & \textbf{0.854} & 43.2\\
                \midrule
        DUSt3R~\citep{wang2024dust3r} & 571M & \multirow{8}{*}{\makecell{Monocular \\ Depth}}& 0.488 & 0.532 & 0.139 & 0.832 & - & - & 1.25 \\
        MASt3R~\citep{leroy2024grounding} & 689M & & 0.413 & 0.569 & 0.123 & 0.833 & - & -  & 1.01\\
        MonST3R~\citep{zhang2025monst3r} & 571M & & 0.402 & 0.525 & 0.069 & 0.954 & - & - & 1.27 \\
        CUT3R~\citep{wang2025continuous} & 793M & & 0.418 & 0.520 & \underline{0.058} & \underline{0.967} & \underline{0.149} & 0.790 & 6.98\\
        Fast3R~\citep{Yang_2025_Fast3R} & 648M & & 0.544 & 0.509 & 0.169 & 0.796 & - & - & 65.8 \\
        FLARE~\citep{zhang2025flare} & 1.40B & & 0.606 & 0.402 & 0.130 & 0.836 & - & - & 1.75\\
        VGGT ~\citep{wang2025vggt} & 1.26B & & \underline{0.292} & \underline{0.629} & 0.071 & 0.947 & 0.160 & \underline{0.799} & 43.2\\
        \textbf{\ourmethod (Ours)} & 1.26B & & \textbf{0.242} & \textbf{0.742} & \textbf{0.053} & \textbf{0.970} & \textbf{0.141} & \textbf{0.840} & 43.2\\
        \bottomrule
    \end{tabular}
    }
    \label{tab:videodepth}
\end{table*}

\section{Experiments}
To evaluate the effectiveness of \ourmethod{}, we apply it to monocular video sequences and assess its performance on five tasks: video depth estimation, monocular depth estimation, camera pose estimation, multi-view point map reconstruction, and 4D view synthesis. We compare against several strong baselines—DUSt3R~\citep{wang2024dust3r}, MASt3R~\citep{leroy2024grounding}, MonST3R~\citep{zhang2025monst3r}, CUT3R~\citep{wang2025continuous}, Fast3R~\citep{Yang_2025_Fast3R}, FLARE~\citep{zhang2025flare}, and VGGT~\citep{wang2025vggt}—across each subtask.

\subsection{Video Depth Estimation}
\begin{figure}[t!]
    \vspace{-0.2cm}
    \includegraphics[width=\linewidth, clip]{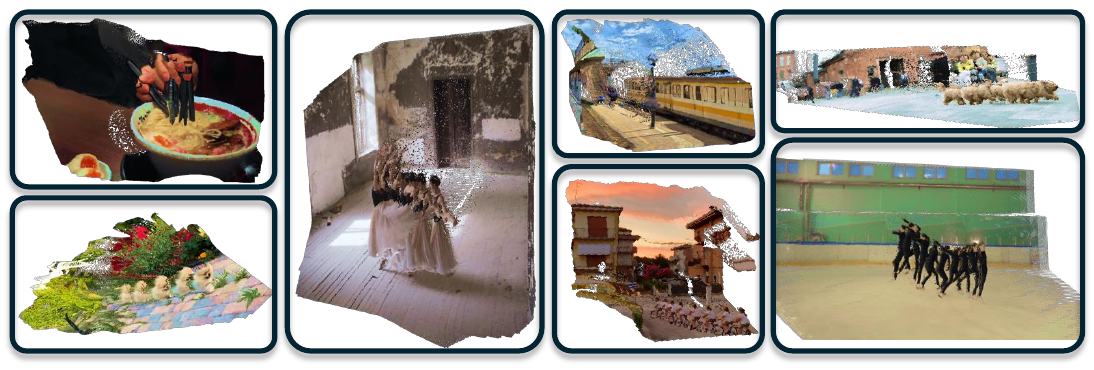}   
    \caption{\textbf{Qualitative Results of Point Cloud Estimation.} \ourmethod{} can estimate camera poses and depth maps from RGB inputs, even in the presence of dynamic objects. (Best viewed in PDF.)}
    \label{fig:point} \vspace{-0.3cm}
\end{figure}

Following the protocol of prior works~\citep{wang2024dust3r, zhang2025monst3r}, we evaluate our approach on the video depth estimation task using Sintel~\citep{Butler:ECCV:2012} and Bonn~\citep{palazzolo2019refusion}. To assess robustness to dynamic objects, we additionally incorporate DyCheck~\citep{gao2022monocular}. We report Absolute Relative Error (Abs Rel) and prediction accuracy at the threshold $\delta < 1.25$, under two alignment settings: (i) scale-only alignment and (ii) joint scale and 3D translation alignment. Qualitative results could be found in Fig~\ref{fig:Depth1_BS} and Fig~\ref{fig:point}.
As summarized in Tab.~\ref{tab:videodepth}, our method establishes a new state of the art across all three datasets and both alignment settings among feed-forward 3D reconstruction models. Compared to VGGT~\citep{wang2025vggt}, which represents the strongest prior baseline, our approach consistently reduces error and improves accuracy. For example, on Sintel with scale-shift alignment, we improve $\delta < 1.25$ accuracy from 0.639 VGGT to 0.763 (+19.4\%) while lowering Abs Rel from 0.261 to 0.212 (-18.8\%). Similar trends are observed on Sintel and Bonn, where our method outperforms VGGT under both alignment regimes, without incurring noticeable increases in speed or memory consumption.

\subsection{Monocular Depth Estimation}
In addition to video depth, we evaluate our approach on monocular depth estimation following \citet{leroy2024grounding, zhou2025manydepth2}. Each predicted depth map is aligned independently with its ground truth, in contrast to the video setting where a single scale (and shift) is applied across the entire sequence. 
As summarized in Tab.~\ref{tab:videodepth}, our method shows consistent improvements over existing feed-forward reconstruction methods. 
In particular, compared to VGGT~\citep{wang2025vggt} in Sintel dataset, our approach reduces Abs Rel from 0.292 to 0.242, and increases $\delta < 1.25$ accuracy from 0.629 to 0.742. While not explicitly optimized for single-frame depth estimation, our method performs favorably against dedicated baselines such as DUSt3R, MONST3R, and FLARE. These results suggest that our model generalizes well from video sequences to single-image inputs.

\begin{wraptable}{r}{0.72\textwidth}  
\vspace{-0pt}                         
\centering
\small
\renewcommand{\arraystretch}{1.1}
\setlength{\tabcolsep}{6pt}
\caption{\textbf{Camera Pose Estimation on Sintel and Tum.}}
    \vspace{-1em}
\resizebox{\linewidth}{!}
{%
\begin{tabular}{lccccccc}
\toprule
\multirow{3}{*}{\textbf{Method}} & \multirow{3}{*}{\textbf{Optim.}} & \multicolumn{3}{c}{\textbf{Sintel}} 
& \multicolumn{3}{c}{\textbf{Tum}} \\

\cmidrule(lr){3-5} \cmidrule(lr){6-8}

& & \cellcolor{col1} ATE $\downarrow$ & \cellcolor{col1} RPE$_{\text{trans}}$ $\downarrow$ & \cellcolor{col1} RPE$_{\text{rot}}$ $\downarrow$ 
& \cellcolor{col1} ATE $\downarrow$ & \cellcolor{col1} RPE$_{\text{trans}}$ $\downarrow$ & \cellcolor{col1} RPE$_{\text{rot}}$ $\downarrow$ \\
\midrule
MonST3R~\citep{zhang2025monst3r}     & $\bullet$     & \textbf{0.108} & \textbf{0.042} & 0.732 & 0.098 & 0.019 & 0.935  \\
DUSt3R~\citep{wang2024dust3r}        &  & 0.417 & 0.250 & 5.796 & 0.140 & 0.106 & 3.286 \\
Spann3R~\citep{wang2024spann3r} &  &  0.329 &  0.110 &  4.471 &  0.056 &  0.021 &  0.591 \\
CUT3R~\citep{wang2025continuous}      &     & 0.213 & 0.066 & 0.621 & 0.046 & 0.015 & 0.473  \\
VGGT~\citep{wang2025vggt}         &         & 0.214 & 0.079 & \underline{0.643} & \underline{0.028} & \underline{0.014} & \underline{0.371}  \\
\textbf{\ourmethod (Ours)}    &                                 & \underline{0.178}    & \underline{0.069}    & \textbf{0.547}    & \textbf{0.016} & \textbf{0.011} & \textbf{0.323}  \\
\bottomrule
\end{tabular}%
}
\label{tab:camera}
\vspace{-4pt}
\end{wraptable}

\subsection{Camera Pose Estimation}
\label{section4.3}
We evaluate camera pose estimation on the dynamic-scene Sintel~\citep{Butler:ECCV:2012} and Tum~\citep{sturm12iros} benchmarks. Following the protocol in~\citep{zhang2025monst3r}, we report Absolute Trajectory Error (ATE), Relative Pose Error in translation (RPE\textsubscript{trans}), and rotation (RPE\textsubscript{rot}). 
For a fair comparison, predicted trajectories are aligned to the ground truth via Sim(3) Umeyama alignment, and we uniformly sample 10 frames per sequence for evaluation.
As shown in Tab.~\ref{tab:camera}, our method delivers substantial improvements on Tum, reducing RPE\textsubscript{trans} by 21\% and RPE\textsubscript{rot} by 13\% compared to prior feed-forward approaches, while maintaining competitive ATE. On Sintel, our approach also reduces  RPE\textsubscript{rot} by 17\%, highlighting its robustness across both synthetic and real-world dynamic scenes.



\subsection{Point Map Estimation}
\label{section4.2}
\begin{wraptable}{r}{0.72\textwidth}
\vspace{-8pt}
\centering
\small
\renewcommand{\arraystretch}{1.1}
\setlength{\tabcolsep}{4pt}
\caption{\textbf{Point reconstruction on DyCheck.}}
\vspace{-1em}
\resizebox{\linewidth}{!}{%
\begin{tabular}{lccccccc}
\toprule
& \multicolumn{7}{c}{\textbf{DyCheck}} \\
\cmidrule(lr){3-8}
\multirow{2}{*}{\textbf{Method}} & \multirow{2}{*}{\textbf{Optim.}}
& \multicolumn{2}{>{\columncolor{col1}}c}{Acc $\downarrow$}
& \multicolumn{2}{>{\columncolor{col1}}c}{Comp $\downarrow$}
& \multicolumn{2}{>{\columncolor{col1}}c}{Overall $\downarrow$} \\
\cmidrule(lr){3-8}
&& Mean & Median & Mean & Median & Mean & Median \\
\midrule
MONST3R~\citep{zhang2025monst3r}    & $\bullet$   & 0.851 & 0.689 & 1.734 & 0.958 & 1.292 & 0.823 \\
DUSt3R~\citep{wang2024dust3r}     &     & 0.802 & 0.595 & 1.950 & 0.815 & 1.376 & 0.705 \\
CUT3R~\citep{wang2025continuous}  &   & \underline{0.458} & \underline{0.342} & 1.633 & \underline{0.792} & \underline{1.042} & 0.567 \\
DAS3R~\cite{xu2024das3r}   & & 1.772 & 1.438 & 2.503 & 1.548 & \textbf{0.475} & \textbf{0.352} \\
VGGT~\citep{wang2025vggt}     &     & 1.051 & 1.016 & \underline{1.594} & 1.393 & 1.322 & 1.204 \\
\textbf{\ourmethod (Ours)}       &      & \textbf{0.403} & \textbf{0.284} & \textbf{1.222} & \textbf{0.728} & 1.115 & \underline{0.559} \\
\bottomrule
\end{tabular}%
}
\label{tab:point}
\vspace{-8pt}
\end{wraptable} 
We further evaluate our method on the DyCheck~\citep{gao2022monocular} benchmark for dynamic-scene point map reconstruction. Following the protocol of~\citep{wang2024dust3r, zhang2025monst3r}, we report Accuracy (Acc.), Completion (Comp.), and Overall error, where lower values indicate better reconstruction quality.
As summarized in Tab.~\ref{tab:point}, our method achieves substantial improvements over prior feed-forward approaches. In particular, compared to the outputs produced by the point head of VGGT~\citep{wang2025vggt}, our approach reduces the mean Accuracy error by more than 60\% (1.051 $\rightarrow$ 0.403) and the median error by over 70\% (1.016 $\rightarrow$ 0.284). Similarly, our method yields consistent gains on Completion, with both mean and median errors reduced by over 20\%.
These results highlight the effectiveness of our dynamic-aware modeling: 
while existing methods either fail to accurately reconstruct moving regions (e.g., Easi3R~\citep{chen2025easi3r}) or show degraded completion under dynamic motion (e.g., MonST3R~\citep{zhang2025monst3r}), 
our method balances accuracy and completeness, producing robust reconstructions in challenging dynamic scenarios.



\subsection{Dynamic Scenes Rendering}
Rendering dynamic scenes has become a key focus in the computer vision community~\citep{wu20244d, pumarola2021d, li2023dynibar, zhou2023dynpoint, zhou2024drivinggaussian}. However, most existing approaches rely heavily on accurate camera poses and high-quality initial point clouds—quantities that are often time-consuming to obtain and particularly challenging to estimate in the presence of complex object motion. \ourmethod{} addresses this limitation by jointly predicting temporally consistent camera poses and dense 3D point clouds directly from RGB sequences containing dynamic content.
To evaluate the utility of \ourmethod{} for dynamic scene rendering, we use its reconstructed point clouds as initialization for the recent 4D-Gaussian splatting framework~\citep{wu20244d} and assess the resulting novel view synthesis quality on the Nerfie benchmark~\citep{gafni2021dynamic}. As shown in Tab.~\ref{tab:nerf_results}, our method consistently achieves superior rendering performance across scenes compared to prior feed-forward 3D reconstruction models. Notably, \ourmethod{} provides a more robust geometric initialization
which leads to improvements over both static-scene baselines (e.g., DUSt3R, VGGT) and recent dynamic-aware methods (e.g., MonST3R, CUT3R), demonstrating its effectiveness as a geometry prior for high-fidelity 4D rendering.

\begin{table*}[t]
\centering
\small
\renewcommand{\arraystretch}{1.1}
\setlength{\tabcolsep}{5pt}
\caption{\textbf{Novel View Synthesis on Nerfie~\citep{gafni2021dynamic}.} 
We report PSNR $\uparrow$, SSIM $\uparrow$, and LPIPS $\downarrow$ for each scene and the average.}
\resizebox{\linewidth}{!}{%
\begin{tabular}{lccccccccccccccccccc}
\toprule
\multirow{2}{*}{\textbf{Method}} 
& \multicolumn{3}{c}{\textbf{chess4}} 
& \multicolumn{3}{c}{\textbf{dvd}} 
& \multicolumn{3}{c}{\textbf{hand8}} 
& \multicolumn{3}{c}{\textbf{laptop8}} 
& \multicolumn{3}{c}{\textbf{tomato-mark8}} 
& \multicolumn{3}{c}{\textbf{Avg}} \\
\cmidrule(lr){2-4} \cmidrule(lr){5-7} \cmidrule(lr){8-10} \cmidrule(lr){11-13} \cmidrule(lr){14-16} \cmidrule(lr){17-19}
& \cellcolor{col2} PSNR $\uparrow$ & \cellcolor{col2} SSIM $\uparrow$ & \cellcolor{col1} LPIPS $\downarrow$ 
& \cellcolor{col2} PSNR $\uparrow$ & \cellcolor{col2} SSIM $\uparrow$ & \cellcolor{col1} LPIPS $\downarrow$
& \cellcolor{col2} PSNR $\uparrow$ & \cellcolor{col2} SSIM $\uparrow$ & \cellcolor{col1} LPIPS $\downarrow$
& \cellcolor{col2} PSNR $\uparrow$ & \cellcolor{col2} SSIM $\uparrow$ & \cellcolor{col1} LPIPS $\downarrow$
& \cellcolor{col2} PSNR $\uparrow$ & \cellcolor{col2} SSIM $\uparrow$ & \cellcolor{col1} LPIPS $\downarrow$
& \cellcolor{col2} PSNR $\uparrow$ & \cellcolor{col2} SSIM $\uparrow$ & \cellcolor{col1} LPIPS $\downarrow$ \\
\midrule
dust3r & 11.572 & 0.263 & 0.633 & 12.363 & 0.494 & 0.546 & 12.445 & 0.269 & 0.639 & 11.818 & 0.185 & 0.622 & 14.961 & 0.361 & 0.564 & 12.632 & 0.314 & 0.601 \\
monst3r & 12.210 & 0.276 & 0.618 & 13.495 & 0.542 & 0.532 & 12.248 & 0.281 & 0.645 & 12.900 & 0.217 & 0.616 & 14.060 & 0.333 & 0.562 & 12.982 & 0.325 & 0.595 \\
cut3r   & \textbf{17.480} & \textbf{0.448} & 0.525 & 14.856 & 0.528 & 0.465 & 14.466 & 0.365 & 0.559 & \textbf{16.505} & \textbf{0.437} & \textbf{0.365} & 17.289 & 0.461 & 0.447 & 16.319 & 0.448 & 0.472 \\
vggt    & 16.807 & 0.390 & \underline{0.497} & \underline{18.288} & \textbf{0.676} & \textbf{0.379} & \underline{15.633} & \underline{0.510} & \underline{0.489} & \underline{15.954} & \underline{0.353} & \underline{0.475} & \underline{17.624} & \underline{0.488} & \underline{0.428} & \underline{16.861} & \underline{0.483} & \underline{0.454} \\
\textbf{\ourmethod (Ours)}  & \underline{17.338} & \underline{0.393} & \textbf{0.491} & \textbf{18.355} & \underline{0.671} & \underline{0.382} & \textbf{18.047} & \textbf{0.536} & \textbf{0.479} & 15.718 & 0.318 & 0.502 & \textbf{18.511} & \textbf{0.504} & \textbf{0.393} & \textbf{17.593} & \textbf{0.485} & \textbf{0.449} \\
\bottomrule
\end{tabular}%
}
\label{tab:nerf_results}
\end{table*}


\subsection{Ablation Studies}
To evaluate the effectiveness of the proposed technique, we perform two ablation studies. First, we examine the fine-tuning strategy by comparing our approach—where only the middle $N_2$ attention layers are updated—with a baseline that fine-tunes all layers of the network (\textsc{VGGT*} (Whole Model)). Second, we study the role of the dynamic-aware aggregator by comparing our full model with a variant that simply fine-tunes the middle $N_2$ layers of VGGT without disentangling dynamics (\textsc{VGGT*} (Middle Layers)). From the comparison between \textsc{VGGT*} (Whole Model) and \textsc{VGGT*} (Middle Layers) as shown in Tab.~\ref{tab:ablation}, we observe that restricting fine-tuning to the middle layers yields performance comparable to full fine-tuning, confirming that these layers capture the most critical information. More importantly, by comparing \textbf{Ours} - \textsc{VGGT*} (Middle Layers + Mask Attention) with \textsc{VGGT*} (Middle Layers), we demonstrate that explicitly disentangling pose and geometry estimation through the dynamic-aware aggregator unlocks the potential of the backbone, leading to substantial performance gains.

\begin{table*}[t]
    \centering
    \caption{
        \textbf{Video Depth Estimation on Sintel~\citep{Butler:ECCV:2012}, Bonn~\citep{palazzolo2019refusion} and DyCheck~\citep{Yang_2025_Fast3R}.} 
    }
    \vspace{-0.4em}
    \resizebox{\linewidth}{!}{%
    \begin{tabular}{lccccccc}
        \toprule
        {\multirow{2}{*}{\textbf{Method}}} &
        {\multirow{2}{*}{\textbf{Align}}} &
        \multicolumn{2}{c}{\textbf{Sintel}} &
        \multicolumn{2}{c}{\textbf{Bonn}} &
        \multicolumn{2}{c}{\textbf{DyCheck}} \\
        \cmidrule(r){3-4} \cmidrule(r){5-6} \cmidrule(r){7-8}
        & & \cellcolor{col1} Abs Rel $\downarrow$ & \cellcolor{col2} $\delta<1.25$ $\uparrow$
          & \cellcolor{col1} Abs Rel $\downarrow$ & \cellcolor{col2} $\delta<1.25$ $\uparrow$
          & \cellcolor{col1} Abs Rel $\downarrow$ & \cellcolor{col2} $\delta<1.25$ $\uparrow$ \\
        \midrule
         \textsc{VGGT*} (Whole Model) & \multirow{3}{*}{\makecell{scale \\ (Video-Depth)}} & \underline{0.405} & \underline{0.593} & 0.101 & \underline{0.891} & \underline{0.175} & \underline{0.775} \\
        \textsc{VGGT*} (Middle Layers) & & 0.409 & 0.590 & \underline{0.099} & 0.879 & 0.177 & 0.766 \\
        \textbf{Ours} - \textsc{VGGT*} (Middle Layers + Mask Attention) & & \textbf{0.357} & \textbf{0.699} & \textbf{0.092} & \textbf{0.904} & \textbf{0.170} & \textbf{0.785} \\
        \bottomrule
    \end{tabular}
    }
    \vspace{-1em}
    \label{tab:ablation}
\end{table*}


\section{Conclusion}

Understanding dynamic scenes remains a central challenge in 4D computer vision, where object motion simultaneously provides valuable geometric cues and disrupts static-scene assumptions critical for camera pose estimation. In this work, we introduce \ourmethod{}, a feedforward framework that adapts a pretrained 3D foundation model to dynamic environments through a disentanglement strategy. Our analysis shows that while VGGT excels in static scenarios, its unified treatment of motion leads to conflicts across tasks. To address this, we propose a dynamics-aware aggregator that disentangles static and dynamic content—suppressing dynamics for pose estimation while leveraging them for geometry and tracking. Combined with a targeted fine-tuning strategy on the most dynamic-sensitive layers, this design unlocks the backbone’s latent capacity for handling motion. Extensive experiments demonstrate that \ourmethod{} achieves state-of-the-art results across depth, pose, and point cloud reconstruction benchmarks. Importantly, we show that effective disentanglement enables strong generalization even with limited dynamic data, paving the way for scalable and efficient 4D scene understanding.


\newpage

\bibliography{iclr2026_conference}

@inproceedings{chefer2021transformer,
  title={Transformer interpretability beyond attention visualization},
  author={Chefer, Hila and Gur, Shir and Wolf, Lior},
  booktitle={Proceedings of the IEEE/CVF conference on computer vision and pattern recognition},
  pages={782--791},
  year={2021}
}

@inproceedings{wang2025vggt,
  title={{VGGT}: Visual geometry grounded transformer},
  author={Wang, Jianyuan and Chen, Minghao and Karaev, Nikita and Vedaldi, Andrea and Rupprecht, Christian and Novotny, David},
  booktitle={Proceedings of the IEEE/CVF Conference on Computer Vision and Pattern Recognition},
  pages={5294--5306},
  year={2025}
}

@article{zhang2025advances,
  title={Advances in Feed-Forward {3D} Reconstruction and View Synthesis: A Survey},
  author={Zhang, Jiahui and Li, Yuelei and Chen, Anpei and Xu, Muyu and Liu, Kunhao and Wang, Jianyuan and Long, Xiao-Xiao and Liang, Hanxue and Xu, Zexiang and Su, Hao and others},
  journal={arXiv preprint arXiv:2507.14501},
  year={2025}
}

@article{zhang2022dino,
  title={Dino: Detr with improved denoising anchor boxes for end-to-end object detection},
  author={Zhang, Hao and Li, Feng and Liu, Shilong and Zhang, Lei and Su, Hang and Zhu, Jun and Ni, Lionel M and Shum, Heung-Yeung},
  journal={arXiv preprint arXiv:2203.03605},
  year={2022}
}

@inproceedings{wang2024dust3r,
  title={{DUSt3R}: Geometric {3D} Vision Made Easy},
  author={Wang, Shuzhe and Leroy, Vincent and Cabon, Yohann and Chidlovskii, Boris and Revaud, Jerome},
  booktitle={Proceedings of the IEEE/CVF Conference on Computer Vision and Pattern Recognition},
  pages={20697--20709},
  year={2024}
}

@inproceedings{zhang2025monst3r,
  title={{MonST3R}: A Simple Approach for Estimating Geometry in the Presence of Motion},
  author={Zhang, Junyi and Herrmann, Charles and Hur, Junhwa and Jampani, Varun and Darrell, Trevor and Cole, Forrester and Sun, Deqing and Yang, Ming-Hsuan},
  booktitle={International Conference on Learning Representations},
  year={2025}
}

@article{han2025d2ust3r,
  title={{D$^2$USt3R}: Enhancing {3D} Reconstruction with {4D} Pointmaps for Dynamic Scenes},
  author={Han, Jisang and An, Honggyu and Jung, Jaewoo and Narihira, Takuya and Seo, Junyoung and Fukuda, Kazumi and Kim, Chaehyun and Hong, Sunghwan and Mitsufuji, Yuki and Kim, Seungryong},
  journal={arXiv preprint arXiv:2504.06264},
  year={2025}
}

@article{chen2025easi3r,
  title={{Easi3r}: Estimating disentangled motion from {DUSt3R} without training},
  author={Chen, Xingyu and Chen, Yue and Xiu, Yuliang and Geiger, Andreas and Chen, Anpei},
  journal={arXiv preprint arXiv:2503.24391},
  year={2025}
}

@inproceedings{li2024megasam,
  title={{MegaSaM}: Accurate, fast and robust structure and motion from casual dynamic videos},
  author={Li, Zhengqi and Tucker, Richard and Cole, Forrester and Wang, Qianqian and Jin, Linyi and Ye, Vickie and Kanazawa, Angjoo and Holynski, Aleksander and Snavely, Noah},
  booktitle={Proceedings of the IEEE/CVF Conference on Computer Vision and Pattern Recognition},
  pages={10486--10496},
  year={2024}
}

@inproceedings{st4rtrack2025,
  title={{St4RTrack}: Simultaneous {4D} Reconstruction and Tracking in the World},
  author={Feng, Haiwen and Zhang, Junyi and Wang, Qianqian and Ye, Yufei and Yu, Pengcheng and Black, Michael J. and Darrell, Trevor and Kanazawa, Angjoo},
  booktitle={Proceedings of the IEEE/CVF International Conference on Computer Vision},
  year={2025}
}

@inproceedings{zheng2023pointodyssey,
  title={{PointOdyssey}: A large-scale synthetic dataset for long-term point tracking},
  author={Zheng, Yang and Harley, Adam W and Shen, Bokui and Wetzstein, Gordon and Guibas, Leonidas J},
  booktitle={Proceedings of the IEEE/CVF International Conference on Computer Vision},
  pages={19855--19865},
  year={2023}
}

@inproceedings{greff2022kubric,
  title={{Kubric}: A scalable dataset generator},
  author={Greff, Klaus and Belletti, Francois and Beyer, Lucas and Doersch, Carl and Du, Yilun and Duckworth, Daniel and Fleet, David J and Gnanapragasam, Dan and Golemo, Florian and Herrmann, Charles and others},
  booktitle={Proceedings of the IEEE/CVF Conference on Computer Vision and Pattern Recognition},
  pages={3749--3761},
  year={2022}
}

@inproceedings{karaev2023dynamicstereo,
  title={{DynamicStereo}: Consistent Dynamic Depth from Stereo Videos},
  author={Karaev, Nikita and Rocco, Ignacio and Graham, Benjamin and Neverova, Natalia and Vedaldi, Andrea and Rupprecht, Christian},
  booktitle={Proceedings of the IEEE/CVF Conference on Computer Vision and Pattern Recognition},
  pages={5858--5868},
  year={2023}
}

@inproceedings{Mehl2023_Spring,
    author    = {Lukas Mehl and Jenny Schmalfuss and Azin Jahedi and Yaroslava Nalivayko and Andr\'{e}s Bruhn},
    title     = {{Spring}: A High-Resolution High-Detail Dataset and Benchmark for Scene Flow, Optical Flow and Stereo},
    booktitle = {Proceedings of the IEEE/CVF Conference on Computer Vision and Pattern Recognition},
    year      = {2023},
    pages     = {4288--4297}
}

@inproceedings{wu20244d,
  title={{4D} Gaussian Splatting for Real-Time Dynamic Scene Rendering},
  author={Wu, Guanjun and Yi, Taoran and Fang, Jiemin and Xie, Lingxi and Zhang, Xiaopeng and Wei, Wei and Liu, Wenyu and Tian, Qi and Wang, Xinggang},
  booktitle={Proceedings of the IEEE/CVF Conference on Computer Vision and Pattern Recognition},
  pages={20310--20320},
  year={2024}
}

@inproceedings{wang2025continuous,
  title={Continuous 3d perception model with persistent state},
  author={Wang, Qianqian and Zhang, Yifei and Holynski, Aleksander and Efros, Alexei A and Kanazawa, Angjoo},
  booktitle={Proceedings of the Computer Vision and Pattern Recognition Conference},
  pages={10510--10522},
  year={2025}
}

@article{luiten2020track,
  title={Track to reconstruct and reconstruct to track},
  author={Luiten, Jonathon and Fischer, Tobias and Leibe, Bastian},
  journal={IEEE Robotics and Automation Letters},
  volume={5},
  number={2},
  pages={1803--1810},
  year={2020},
  publisher={IEEE}
}

@inproceedings{gafni2021dynamic,
  title={Dynamic neural radiance fields for monocular 4d facial avatar reconstruction},
  author={Gafni, Guy and Thies, Justus and Zollhofer, Michael and Nie{\ss}ner, Matthias},
  booktitle={Proceedings of the IEEE/CVF Conference on Computer Vision and Pattern Recognition},
  pages={8649--8658},
  year={2021}
}

@inproceedings{zhang2025flare,
  title={{FLARE}: Feed-forward geometry, appearance and camera estimation from uncalibrated sparse views},
  author={Zhang, Shangzhan and Wang, Jianyuan and Xu, Yinghao and Xue, Nan and Rupprecht, Christian and Zhou, Xiaowei and Shen, Yujun and Wetzstein, Gordon},
  booktitle={Proceedings of the IEEE/CVF Conference on Computer Vision and Pattern Recognition},
  pages={21936--21947},
  year={2025}
}

@article{bochkovskii2024depth,
  title={Depth pro: Sharp monocular metric depth in less than a second},
  author={Bochkovskii, Aleksei and Delaunoy, Ama{\~A}{\c{G}}l and Germain, Hugo and Santos, Marcel and Zhou, Yichao and Richter, Stephan R and Koltun, Vladlen},
  journal={arXiv preprint arXiv:2410.02073},
  year={2024}
}

@inproceedings{kopf2021robust,
  title={Robust consistent video depth estimation},
  author={Kopf, Johannes and Rong, Xuejian and Huang, Jia-Bin},
  booktitle={Proceedings of the IEEE/CVF Conference on Computer Vision and Pattern Recognition},
  pages={1611--1621},
  year={2021}
}

@inproceedings{mustafa2016temporally,
  title={Temporally coherent {4D} reconstruction of complex dynamic scenes},
  author={Mustafa, Armin and Kim, Hansung and Guillemaut, Jean-Yves and Hilton, Adrian},
  booktitle={Proceedings of the IEEE Conference on Computer Vision and Pattern Recognition},
  pages={4660--4669},
  year={2016}
}

@article{lin2025movies,
  title={{MoVieS}: Motion-Aware {4D} Dynamic View Synthesis in One Second},
  author={Lin, Chenguo and Lin, Yuchen and Pan, Panwang and Yu, Yifan and Yan, Honglei and Fragkiadaki, Katerina and Mu, Yadong},
  journal={arXiv preprint arXiv:2507.10065},
  year={2025}
}

@article{zhuo2025streamvggt,
  title={Streaming {4D} Visual Geometry Transformer},
  author={Zhuo, Dong and Zheng, Wenzhao and Guo, Jiahe and Wu, Yuqi and Zhou, Jie and Lu, Jiwen},
  journal={arXiv preprint arXiv:2507.11539},
  year={2025}
}

@article{gao2022monocular,
  title={Monocular dynamic view synthesis: A reality check},
  author={Gao, Hang and Li, Ruilong and Tulsiani, Shubham and Russell, Bryan and Kanazawa, Angjoo},
  journal={Advances in Neural Information Processing Systems},
  volume={35},
  pages={33768--33780},
  year={2022}
}

@inproceedings{Butler:ECCV:2012,
  title = {A naturalistic open source movie for optical flow evaluation},
  author = {Butler, D. J. and Wulff, J. and Stanley, G. B. and Black, M. J.},
  booktitle = {European Conference on Computer Vision},
  pages = {611--625},
  year = {2012},
  publisher = {Springer}
}

@article{tang2024mvdust3rsinglestagescenereconstruction,
  title={{MV-DUSt3R+}: Single-Stage Scene Reconstruction from Sparse Views In 2 Seconds}, 
  author={Zhenggang Tang and Yuchen Fan and Dilin Wang and Hongyu Xu and Rakesh Ranjan and Alexander Schwing and Zhicheng Yan},
  journal={arXiv preprint arXiv:2412.06974},
  year={2024}
}

@inproceedings{Yang_2025_Fast3R,
  title={{Fast3R}: Towards {3D} Reconstruction of 1000+ Images in One Forward Pass},
  author={Jianing Yang and Alexander Sax and Kevin J. Liang and Mikael Henaff and Hao Tang and Ang Cao and Joyce Chai and Franziska Meier and Matt Feiszli},
  booktitle={Proceedings of the IEEE/CVF Conference on Computer Vision and Pattern Recognition},
  year={2025}
}

@inproceedings{reizenstein21co3d,
  author = {Reizenstein, Jeremy and Shapovalov, Roman and Henzler, Philipp and Sbordone, Luca and Labatut, Patrick and Novotny, David},
  title = {Common Objects in {3D}: Large-Scale Learning and Evaluation of Real-life {3D} Category Reconstruction},
  booktitle = {Proceedings of the IEEE/CVF International Conference on Computer Vision},
  year = {2021}
}

@inproceedings{pumarola2021d,
  title={{D-NeRF}: Neural radiance fields for dynamic scenes},
  author={Pumarola, Albert and Corona, Enric and Pons-Moll, Gerard and Moreno-Noguer, Francesc},
  booktitle={Proceedings of the IEEE/CVF Conference on Computer Vision and Pattern Recognition},
  pages={10318--10327},
  year={2021}
}

@inproceedings{li2023dynibar,
  title={{DynIBaR}: Neural dynamic image-based rendering},
  author={Li, Zhengqi and Wang, Qianqian and Cole, Forrester and Tucker, Richard and Snavely, Noah},
  booktitle={Proceedings of the IEEE/CVF Conference on Computer Vision and Pattern Recognition},
  pages={4273--4284},
  year={2023}
}

@article{zhou2023dynpoint,
  title={{DynPoint}: Dynamic neural point for view synthesis},
  author={Zhou, Kaichen and Zhong, Jia-Xing and Shin, Sangyun and Lu, Kai and Yang, Yiyuan and Markham, Andrew and Trigoni, Niki},
  journal={Advances in Neural Information Processing Systems},
  volume={36},
  pages={69532--69545},
  year={2023}
}

@inproceedings{zhou2024drivinggaussian,
  title={{DrivingGaussian}: Composite Gaussian splatting for surrounding dynamic autonomous driving scenes},
  author={Zhou, Xiaoyu and Lin, Zhiwei and Shan, Xiaojun and Wang, Yongtao and Sun, Deqing and Yang, Ming-Hsuan},
  booktitle={Proceedings of the IEEE/CVF Conference on Computer Vision and Pattern Recognition},
  pages={21634--21643},
  year={2024}
}

@inproceedings{zhang2022structure,
  title={Structure and motion from casual videos},
  author={Zhang, Zhoutong and Cole, Forrester and Li, Zhengqi and Rubinstein, Michael and Snavely, Noah and Freeman, William T},
  booktitle={European Conference on Computer Vision},
  pages={20--37},
  year={2022},
  publisher={Springer}
}

@inproceedings{yin2023metric3d,
  title={{Metric3D}: Towards zero-shot metric {3D} prediction from a single image},
  author={Yin, Wei and Zhang, Chi and Chen, Hao and Cai, Zhipeng and Yu, Gang and Wang, Kaixuan and Chen, Xiaozhi and Shen, Chunhua},
  booktitle={Proceedings of the IEEE/CVF International Conference on Computer Vision},
  pages={9043--9053},
  year={2023}
}

@article{bhat2023zoedepthzeroshottransfercombining,
  title={{ZoeDepth}: Zero-shot Transfer by Combining Relative and Metric Depth}, 
  author={Shariq Farooq Bhat and Reiner Birkl and Diana Wofk and Peter Wonka and Matthias M{\"u}ller},
  journal={arXiv preprint arXiv:2302.12288},
  year={2023}
}

@article{tang2019banetdensebundleadjustment,
  title={{BA-Net}: Dense Bundle Adjustment Network}, 
  author={Chengzhou Tang and Ping Tan},
  journal={arXiv preprint arXiv:1806.04807},
  year={2018}
}

@inproceedings{Yao_2018_ECCV,
  author = {Yao, Yao and Luo, Zixin and Li, Shiwei and Fang, Tian and Quan, Long},
  title = {{MVSNet}: Depth Inference for Unstructured Multi-view Stereo},
  booktitle = {European Conference on Computer Vision},
  year = {2018}
}

@article{chen2021mvsnerf,
  title={{MVSNeRF}: Fast Generalizable Radiance Field Reconstruction from Multi-View Stereo},
  author={Chen, Anpei and Xu, Zexiang and Zhao, Fuqiang and Zhang, Xiaoshuai and Xiang, Fanbo and Yu, Jingyi and Su, Hao},
  journal={arXiv preprint arXiv:2103.15595},
  year={2021}
}

@article{lu2024align3r,
  title={{Align3r}: Aligned monocular depth estimation for dynamic videos},
  author={Lu, Jiahao and Huang, Tianyu and Li, Peng and Dou, Zhiyang and Lin, Cheng and Cui, Zhiming and Dong, Zhen and Yeung, Sai-Kit and Wang, Wenping and Liu, Yuan},
  journal={arXiv preprint arXiv:2412.03079},
  year={2024}
}

@article{Oliensis2000,
    author  = {Oliensis, John},
    title   = {A Critique of Structure-from-Motion Algorithms},
    journal = {Computer Vision and Image Understanding},
    volume  = {80},
    number  = {2},
    pages   = {172--214},
    year    = {2000},
    publisher = {Elsevier}
}

@article{ozyesil2017surveystructuremotion,
  title={A Survey of Structure from Motion}, 
  author={Onur Ozyesil and Vladislav Voroninski and Ronen Basri and Amit Singer},
  journal={arXiv preprint arXiv:1701.08493},
  year={2017}
}

@article{point3r,
  title={{Point3R}: Streaming {3D} Reconstruction with Explicit Spatial Pointer Memory}, 
  author={Yuqi Wu and Wenzhao Zheng and Jie Zhou and Jiwen Lu},
  journal={arXiv preprint arXiv:2507.02863},
  year={2025}
}

@article{wang2024spann3r,
  title={{3D} Reconstruction with Spatial Memory},
  author={Wang, Hengyi and Agapito, Lourdes},
  journal={arXiv preprint arXiv:2408.16061},
  year={2024}
}

@article{xu2025geometrycrafterconsistentgeometryestimation,
  title={{GeometryCrafter}: Consistent Geometry Estimation for Open-world Videos with Diffusion Priors}, 
  author={Tian-Xing Xu and Xiangjun Gao and Wenbo Hu and Xiaoyu Li and Song-Hai Zhang and Ying Shan},
  journal={arXiv preprint arXiv:2504.01016},
  year={2025}
}

@article{jiang2025geo4d,
  title={{Geo4D}: Leveraging Video Generators for Geometric {4D} Scene Reconstruction}, 
  author={Zeren Jiang and Chuanxia Zheng and Iro Laina and Diane Larlus and Andrea Vedaldi},
  journal={arXiv preprint arXiv:2504.07961},
  year={2025}
}

@inproceedings{Yao_2025_CVPR,
    author    = {Yao, David Yifan and Zhai, Albert J. and Wang, Shenlong},
    title     = {{Uni4D}: Unifying Visual Foundation Models for {4D} Modeling from a Single Video},
    booktitle = {Proceedings of the IEEE/CVF Conference on Computer Vision and Pattern Recognition},
    pages     = {1116--1126},
    year      = {2025}
}

@inproceedings{jin2025stereo4d,
    author    = {Jin, Linyi and Tucker, Richard and Li, Zhengqi and Fouhey, David and Snavely, Noah and Holynski, Aleksander},
    title     = {{Stereo4D}: Learning How Things Move in {3D} from Internet Stereo Videos},
    booktitle = {Proceedings of the IEEE/CVF Conference on Computer Vision and Pattern Recognition},
    year      = {2025}
}

@inproceedings{piccinelli2024unidepth,
    author    = {Piccinelli, Luigi and Yang, Yung-Hsu and Sakaridis, Christos and Segu, Mattia and Li, Siyuan and Van Gool, Luc and Yu, Fisher},
    title     = {{UniDepth}: Universal Monocular Metric Depth Estimation},
    booktitle = {Proceedings of the IEEE/CVF Conference on Computer Vision and Pattern Recognition},
    pages     = {10106--10116},
    year      = {2024}
}

@article{cao2025reconstructing,
  title={Reconstructing {4D} Spatial Intelligence: A Survey},
  author={Cao, Yukang and Lu, Jiahao and Huang, Zhisheng and Shen, Zhuowei and Zhao, Chengfeng and Hong, Fangzhou and Chen, Zhaoxi and Li, Xin and Wang, Wenping and Liu, Yuan and others},
  journal={arXiv preprint arXiv:2507.21045},
  year={2025}
}

@inproceedings{tian2023mononerf,
  title={{MonoNeRF}: Learning a generalizable dynamic radiance field from monocular videos},
  author={Tian, Fengrui and Du, Shaoyi and Duan, Yueqi},
  booktitle={Proceedings of the IEEE/CVF International Conference on Computer Vision},
  pages={17903--17913},
  year={2023}
}

@inproceedings{van2022revealing,
  title={Revealing occlusions with {4D} neural fields},
  author={Van Hoorick, Basile and Tendulkar, Purva and Sur{\'\i}s, D{\'\i}dac and Park, Dennis and Stent, Simon and Vondrick, Carl},
  booktitle={Proceedings of the IEEE/CVF Conference on Computer Vision and Pattern Recognition},
  pages={3011--3021},
  year={2022}
}

@inproceedings{busching2024flowibr,
  title={{FlowIBR}: Leveraging pre-training for efficient neural image-based rendering of dynamic scenes},
  author={B{\"u}sching, Marcel and Bengtson, Josef and Nilsson, David and Bj{\"o}rkman, M{\aa}rten},
  booktitle={Proceedings of the IEEE/CVF Conference on Computer Vision and Pattern Recognition},
  pages={8016--8026},
  year={2024}
}

@article{liang2024feed,
  title={Feed-forward bullet-time reconstruction of dynamic scenes from monocular videos},
  author={Liang, Hanxue and Ren, Jiawei and Mirzaei, Ashkan and Torralba, Antonio and Liu, Ziwei and Gilitschenski, Igor and Fidler, Sanja and Oztireli, Cengiz and Ling, Huan and Gojcic, Zan and others},
  journal={arXiv preprint arXiv:2412.03526},
  year={2024}
}

@article{li2025light,
  title={Light of Normals: Unified Feature Representation for Universal Photometric Stereo},
  author={Li, Hong and Chen, Houyuan and Ye, Chongjie and Chen, Zhaoxi and Li, Bohan and Xu, Shaocong and Guo, Xianda and Liu, Xuhui and Wang, Yikai and Zhang, Baochang and others},
  journal={arXiv preprint arXiv:2506.18882},
  year={2025}
}

@article{zhao2023pseudo,
  title={Pseudo-generalized dynamic view synthesis from a video},
  author={Zhao, Xiaoming and Colburn, Alex and Ma, Fangchang and Bautista, Miguel Angel and Susskind, Joshua M and Schwing, Alexander G},
  journal={arXiv preprint arXiv:2310.08587},
  year={2023}
}

@inproceedings{sun2020scalability,
  title={Scalability in perception for autonomous driving: Waymo open dataset},
  author={Sun, Pei and Kretzschmar, Henrik and Dotiwalla, Xerxes and Chouard, Aurelien and Patnaik, Vijaysai and Tsui, Paul and Guo, James and Zhou, Yin and Chai, Yuning and Caine, Benjamin and others},
  booktitle={Proceedings of the IEEE/CVF conference on computer vision and pattern recognition},
  pages={2446--2454},
  year={2020}
}

@inproceedings{leroy2024grounding,
  title={Grounding image matching in 3d with mast3r},
  author={Leroy, Vincent and Cabon, Yohann and Revaud, J{\'e}r{\^o}me},
  booktitle={European Conference on Computer Vision},
  pages={71--91},
  year={2024},
  organization={Springer}
}

@inproceedings{palazzolo2019refusion,
  title={ReFusion: 3D reconstruction in dynamic environments for RGB-D cameras exploiting residuals},
  author={Palazzolo, Emanuele and Behley, Jens and Lottes, Philipp and Giguere, Philippe and Stachniss, Cyrill},
  booktitle={2019 IEEE/RSJ International Conference on Intelligent Robots and Systems (IROS)},
  pages={7855--7862},
  year={2019},
  organization={IEEE}
}

@article{zhou2025manydepth2,
  title={Manydepth2: Motion-aware self-supervised monocular depth estimation in dynamic scenes},
  author={Zhou, Kaichen and Bian, Jia-Wang and Zheng, Jian-Qing and Zhong, Jiaxing and Xie, Qian and Trigoni, Niki and Markham, Andrew},
  journal={IEEE Robotics and Automation Letters},
  year={2025},
  publisher={IEEE}
}

@article{raghu2021vision,
  title={Do vision transformers see like convolutional neural networks?},
  author={Raghu, Maithra and Unterthiner, Thomas and Kornblith, Simon and Zhang, Chiyuan and Dosovitskiy, Alexey},
  journal={Advances in neural information processing systems},
  volume={34},
  pages={12116--12128},
  year={2021}
}

@article{xu2024das3r,
  title={Das3r: Dynamics-aware gaussian splatting for static scene reconstruction},
  author={Xu, Kai and Tse, Tze Ho Elden and Peng, Jizong and Yao, Angela},
  journal={arXiv preprint arXiv:2412.19584},
  year={2024}
}

@inproceedings{caron2021emerging,
  title={Emerging properties in self-supervised vision transformers},
  author={Caron, Mathilde and Touvron, Hugo and Misra, Ishan and J{\'e}gou, Herv{\'e} and Mairal, Julien and Bojanowski, Piotr and Joulin, Armand},
  booktitle={Proceedings of the IEEE/CVF international conference on computer vision},
  pages={9650--9660},
  year={2021}
}

@inproceedings{sturm12iros,
 author = {J. Sturm and N. Engelhard and F. Endres and W. Burgard and D. Cremers},
 title = {A Benchmark for the Evaluation of RGB-D SLAM Systems},
 booktitle = {Proc. of the International Conference on Intelligent Robot Systems (IROS)},
 year = {2012},
 month = {Oct.},
}
\bibliographystyle{iclr2026_conference}

\newpage
\section*{Appendix}

\section{Qualitative Results}
\begin{figure}[h]
\vspace{-1.em}
    \includegraphics[width=\textwidth]{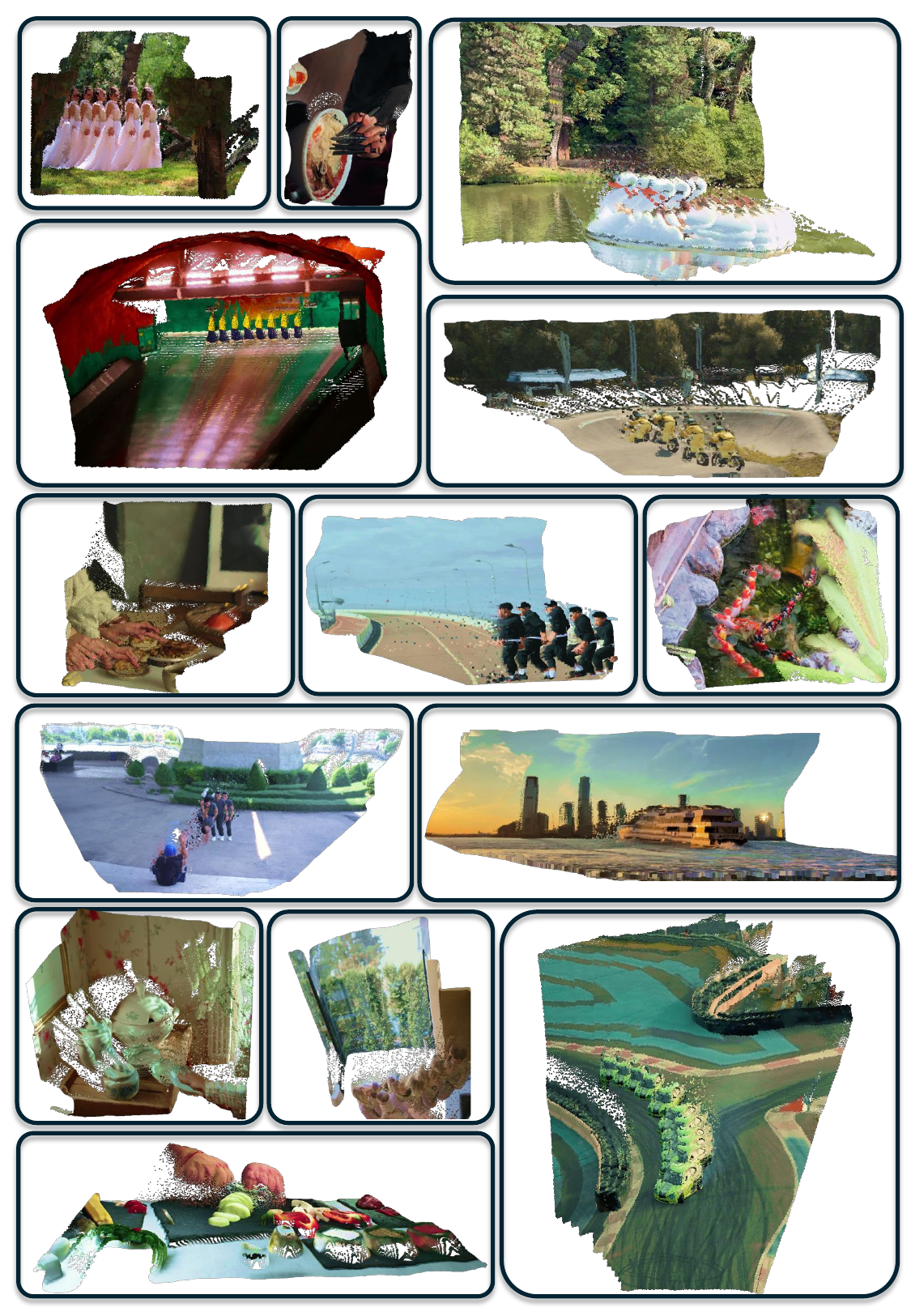}
    \caption{\textbf{\ourmethod} takes a sequence of RGB images depicting a dynamic scene as input and simultaneously predicts the corresponding camera parameters and 3D geometry information—all within a fraction of a second. (Best viewed in PDF.) 
    }
    \label{fig:appendix}
\end{figure}

To further assess the generalization ability of \ourmethod{}, we apply it to a diverse set of in-the-wild video sequences, as illustrated in Fig.~\ref{fig:appendix}. These examples span a wide range of dynamic scenarios, including human activities, object interactions, and complex background motions. We observe that \ourmethod{} consistently produces stable and accurate predictions across all cases, effectively handling both static and highly dynamic regions. Notably, the method demonstrates strong robustness even under challenging conditions such as occlusions, fast motion, and varying illumination, highlighting its applicability beyond controlled benchmarks and into real-world video data.

\section{Acknowledgment}
All video materials used in this study were obtained from Pexels (https://www.pexels.com) and are distributed under the free Pexels License. While the license does not require individual attribution, we would like to acknowledge and thank the Pexels creators whose work contributed to the preparation of our figures and demonstrations. 

\section{Architecture Design}

\subsection{Choice of Middle-Layer Fine-Tuning Strategy}

To better understand the computational characteristics of our baseline, we analyze the parameter distribution of VGGT~\citep{wang2025vggt}, as summarized in Tab.~\ref{tab:size}. The majority of parameters are concentrated in the global attention blocks, which dominate both representational capacity and memory footprint. In contrast, the camera, depth, and point heads account for only a small fraction of the parameters, suggesting that most of the network capacity is devoted to learning general-purpose spatiotemporal features rather than task-specific decoding.
\begin{wraptable}{r}{0.62\textwidth} 
\vspace{-0pt}
\centering
\small
\renewcommand{\arraystretch}{1.1}
\setlength{\tabcolsep}{4pt}
\caption{\textbf{Parameter distribution across modules.} "M" denotes millions of parameters.}
\vspace{-1em}
\resizebox{\linewidth}{!}{%
\begin{tabular}{lccccc}
\toprule
\textbf{Module} & Depth & Point & Track & Camera & Aggregator \\
\midrule
\textbf{Parameters} & 32.7M & 32.7M & 65.9M & 216.2M & 909.1M \\
\bottomrule
\end{tabular}%
}
\label{tab:size}
\vspace{-6pt}
\end{wraptable}
This observation leads to two key insights. First, fine-tuning the entire VGGT backbone is computationally inefficient: updating all attention layers substantially increases runtime and storage costs without yielding proportional gains. Second, since most parameters reside in global attention, selectively adapting the layers most sensitive to dynamic content is a more effective way to exploit the backbone’s capacity.

\begin{wraptable}{r}{0.62\textwidth} 
\vspace{-1.3em}
\centering
\caption{\textbf{Study of Different Masking Strategies Applied to VGGT.}
This experiment is conducted on the Odyssey dataset. We evaluate unscaled pose estimation using Relative Translation Error (RPE
trans) and Relative Rotation Error (RPE rot). For static regions, Static-D denotes the Absolute Depth Error, and Static-T represents the Average Endpoint Error (EPE) for 2D point tracking. 
}
\resizebox{\linewidth}{!}{
\begin{tabular}{lcccc}
\toprule
\textbf{Method} & \cellcolor{col1}\textbf{RPE trans} ↓ & \cellcolor{col1}\textbf{RPE rot} ↓ & \cellcolor{col1}\textbf{Static-D} ↓ & \cellcolor{col1}\textbf{Static-T} ↓ \\
\midrule
Normal            & 0.244 & 0.942 & 0.085 & 17.071 \\
Input-MSK        & 0.246 & 1.006 & 0.099 & 17.866 \\
DD-MSK            & 0.243 & 0.869 & 0.566 & 24.797 \\
w/o 4th       & - & - & 0.114 & 18.821 \\
w/o 11th        & - & - & 0.095 & 18.403 \\
w/o 17th       & - & - & 1.663 & 39.841 \\
w/o 23rd       & - & - & 0.103 & 17.645 \\
\bottomrule
\end{tabular}}
\label{tab:initial_res}
\end{wraptable}

To probe VGGT’s handling of dynamic regions, we examine the role of attention maps within the global attention blocks—specifically at layers 4, 11, 17, and 23—which also provide inputs to the geometry decoder. To assess their contribution, we sequentially replace each layer’s output with random noise and report the results in Tab.~\ref{tab:initial_res}. Since only the last layer is fed into the camera head, randomizing other layers does not affect camera estimation performance; therefore, we omit camera estimation results here.
The results show that the 17\textsuperscript{th} layer exerts the strongest influence on geometry quality, underscoring the non-uniform importance of layers.
Motivated by these findings, we propose a targeted fine-tuning strategy: adapting only the middle 10 attention layers, which are most responsive to dynamic content. This achieves performance comparable to or better than full fine-tuning, while substantially reducing computational overhead. 

During training, we observe that keeping the dynamic mask throughout the entire optimization process, or applying the mask only in the first stage and removing it in the second stage, yields similar performance. Notably, both strategies consistently outperform the baseline model trained with the original backbone without masking. Therefore, in our implementation, we provide both variants.

\subsection{Design of Pose–Geometry Disentanglement}

Figure~\ref{fig:motivation}(b) shows that dynamic objects consistently receive lower attention weights, indicating the model learns to suppress them rather than incorporate their motion. We therefore test baseline strategies inspired by prior work on motion-aware modeling \cite{chen2025easi3r}:  
\begin{itemize}
    \item \underline{Input-MSK (Input masking):} Mask out dynamic regions at the image input level.  
    \item \underline{DD-MSK (Dynamic–Dynamic masking):} Suppress attention among dynamic tokens themselves.  
\end{itemize}

Specifically, the DD-MSK strategy aims to mitigate the instability caused by excessive interactions within dynamic regions. Let $\mathcal{M}_{\text{Dynamic}} \in \mathbb{R}^{B \times (S \cdot P) \times C}$ denote the dynamic feature mask, where all patch tokens corresponding to dynamic regions are preserved. The DD-MSK is then constructed as:
\begin{equation}
\mathcal{M}_{\text{DD-MSK}} = \mathcal{M}_{\text{Dynamic}} \cdot \mathcal{M}_{\text{Dynamic}}^\top,
\end{equation}
which blocks self-attention among dynamic patches while still allowing them to attend to static ones. In practice, this prevents dynamic regions from reinforcing noisy patterns, leading to more stable pose estimation and geometry reconstruction.

As shown in Table~\ref{tab:initial_res}, DD-MSK improves pose estimation by isolating reliable motion cues, but simultaneously degrades geometry estimation, which requires integrating both static and dynamic motion signals. This observation aligns with our architectural design: dynamic information should be disentangled across tasks rather than globally suppressed.

\noindent\textbf{Summary.} Our structural and empirical analysis supports two conclusions:  
\textbf{1. Middle attention layers are more influential.} Perturbation analysis highlights the dominance of deeper global attention layers, particularly the 17\textsuperscript{th}, motivating fine-tuning only the middle 10 layers.  
\textbf{2. Rigid masking is suboptimal.} Suppressing dynamic patches improves pose estimation but harms geometry reconstruction, confirming the need for task-specific disentanglement.

\begin{figure*}[t!]
    \vspace{-1.8em}
    \includegraphics[width=\linewidth, clip]{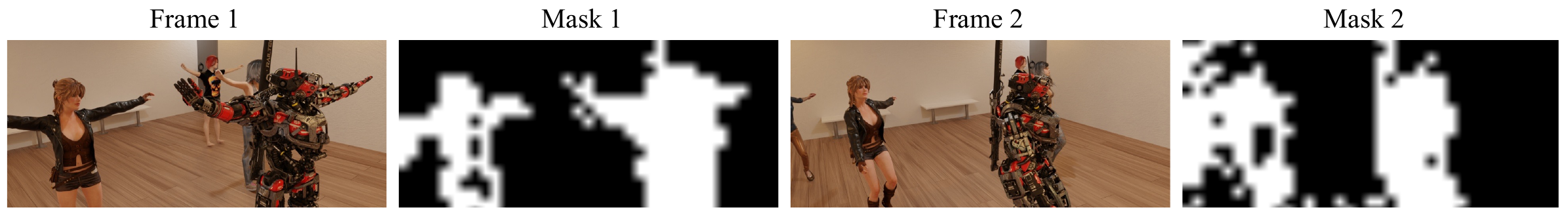}   
    \vspace{-0.7cm}
    \caption{\textbf{Visualization of learned masks.} Our dynamic mask prediction module effectively captures dynamic content in the scene without explicit supervision. (Best viewed in PDF.)}
    \label{fig:Mask_vis} \vspace{-0.3cm}
\end{figure*}

\section{Training Details}

We train our model on a diverse mixture of dynamic and static datasets, including Odyssey, DynamicReplica, Kubric-MV, Spring, CO3D, Waymo, Sintel and our internal dataset, with a total of approximately 2.39M sampled sequences as in Tab.~\ref{tab:data}. To balance domain diversity, we assign per-dataset sampling multipliers and cap the number of training batches per epoch. Images are resized to $518 \times 518$ with patch size 14. We adopt AdamW with an initial learning rate of $1\times 10^{-5}$, weight decay $0.01$, and gradient clipping of 1.0. Mixed precision (bfloat16) training is used to improve efficiency. Following our middle-layer fine-tuning strategy, we freeze the shallow and final blocks of VGGT while updating only the middle 10 global attention layers, which we identify as most sensitive to dynamic content. The multitask loss combines camera, depth, and point supervision with weights of 5.0, 1.0, and 1.0, respectively. 

\begin{wraptable}{r}{0.62\textwidth} 
\centering
\caption{\textbf{Datasets used in the fine-tuning process.} 
\textbf{Dynamic} indicates whether the dataset contains dynamic objects. 
\textbf{Frames} and \textbf{Scenes} denote the number of image frames and unique object-centric scenes. 
\textbf{Ratio} is the scene-level sampling multiplier used to balance datasets during training.}
\resizebox{\linewidth}{!}{
\begin{tabular}{lccccc}
\toprule
\textbf{Dataset} & \textbf{Dynamic} & \textbf{Realistic} & \textbf{Frames} & \textbf{Scenes} & \textbf{Ratio} \\
\midrule
CO3D~\cite{reizenstein21co3d} & $\times$ & $\checkmark$ & 1.5M  & 19K   & 20\% \\
PointOdyssey~\cite{zheng2023pointodyssey} & $\checkmark$ & $\times$ & 6K   & 131   & 10\% \\
Kubric-MV~\cite{greff2022kubric} & $\checkmark$ & $\times$ & 70K  & 3K    & 10\% \\
DynamicReplica~\cite{karaev2023dynamicstereo} & $\checkmark$ & $\times$ & 145K & 484   & 20\% \\
Spring~\cite{Mehl2023_Spring} & $\checkmark$ & $\times$ & 200K & 37    & 10\% \\
Waymo~\cite{sun2020scalability} & $\checkmark$ & $\checkmark$ & 230K & 1.1K  & 10\% \\
Ours & $\checkmark$ & $\checkmark$ & 480K & 2K  & 20\% \\
\bottomrule
\end{tabular}}
\label{tab:data}
\end{wraptable}

\section{Architecture Analysis}
To further evaluate the effectiveness of our dynamic mask prediction module, we visualize the predicted masks on sequences from the Odyssey dataset. As shown in Fig.~\ref{fig:Mask_vis}, the learned masks successfully highlight moving objects such as people and vehicles, while leaving static backgrounds largely unmarked. Importantly, this separation emerges without any explicit supervision, indicating that the model is able to infer dynamic regions purely from motion cues and spatiotemporal inconsistencies. This validates our design choice: the dynamic mask prediction module provides a reliable mechanism to disentangle dynamic and static content, thereby improving the robustness of pose estimation and geometry reconstruction in challenging dynamic scenes.

\newpage

\end{document}